\documentclass[11pt,letterpaper]{article}

\PassOptionsToPackage{table}{xcolor}

\usepackage{amsmath,amssymb,amsthm,amsfonts}
\usepackage{bm}
\usepackage{deepthink}
\titleformat{\section}
  {\Large\sffamily\bfseries}
  {\color{DTAccent}\thesection}{0.8em}{\color{DTAccent}}
\titleformat{\subsection}
  {\large\sffamily\bfseries}
  {\color{DTAccent}\thesubsection}{0.8em}{\color{DTAccent}}
\usepackage{hyperref}
\usepackage{url}
\usepackage{booktabs}
\usepackage{microtype}
\usepackage{xcolor}
\usepackage{graphicx}
\usepackage{algorithm}
\usepackage{algorithmic}
\usepackage{placeins}
\usepackage[nameinlink]{cleveref}
\usepackage[
  backend=biber,
  style=alphabetic,
  natbib=true,
  maxbibnames=100,
  minbibnames=100,
  maxcitenames=2,
  mincitenames=2
]{biblatex}
\usepackage[most]{tcolorbox}
\usepackage{subcaption}
\usepackage{titletoc}
\usepackage{enumitem}

\newtheorem{proposition}{Proposition}

\crefname{equation}{Eq.}{Eqs.}
\Crefname{equation}{Equation}{Equations}
\crefname{figure}{Fig.}{Figs.}
\Crefname{figure}{Figure}{Figures}
\crefname{table}{Table}{Tables}
\Crefname{table}{Table}{Tables}
\crefname{section}{Sec.}{Secs.}
\Crefname{section}{Section}{Sections}
\crefname{appendix}{App.}{App.}
\Crefname{appendix}{Appendix}{Appendices}
\crefname{proposition}{Prop.}{Props.}
\Crefname{proposition}{Proposition}{Propositions}

\newcommand{\DAWNcomment}[1]{\hfill$\triangleright$~\textit{#1}}

\newcommand{\best}[1]{\textbf{#1}}

\title{DAWN: Robust Neural Blind Deconvolution\\ via Alternating Wiener Guidance}

\newcommand{\jointlast}{\textsuperscript{\dag}}

\authorblock{
  \href{https://yidapan.org/}{\textbf{Yida Pan}}\textsuperscript{1},
  \href{https://umjiayx.github.io/}{\textbf{Yixuan Jia}}\textsuperscript{1},
  \href{https://huang-chao-yan.github.io/}{\textbf{Chaoyan Huang}}\textsuperscript{2,1},
  \href{https://qingqu.engin.umich.edu/}{\textbf{Qing Qu}}\textsuperscript{1,}\jointlast,
  \href{https://sites.google.com/view/ismailalkhouri/about?authuser=0}{\textbf{Ismail R. Alkhouri}}\textsuperscript{3,1,}\jointlast
}

\affiliation{
  \textsuperscript{1}University of Michigan \quad $\cdot$ \quad
  \textsuperscript{2}Michigan State University \quad $\cdot$ \quad
  \textsuperscript{3}Los Alamos National Laboratory
}

\authornote{
  \jointlast\ Joint last author
}

\abstracttext{
Blind image deconvolution (BID) is an ill-posed inverse problem that aims to jointly recover a sharp image and blur kernel from a single degraded observation. Under nontrivial and high noise, which is common in photography and scientific imaging such as microscopy, BID remains particularly challenging, as existing methods suffer from noise overfitting and inaccurate kernel estimation. Existing training-data-free learning methods for BID, such as those based on the deep image prior (DIP), often utilize external guidance by including model-based estimates as additional loss terms, which underexploits the information these estimates carry especially under nontrivial noise. To address these challenges, we propose DAWN, a training-data-free DIP framework that alternates between injecting model-based estimates directly into the model input and refining this imperfect estimate with the network's own prior. To provide more reliable kernel estimates for these injections, we apply autoencoding regularization to both the image and kernel networks. On natural and scientific images under varying levels of Gaussian and Poisson noise, DAWN outperforms state-of-the-art methods in both image and kernel quality (achieving up to +4.5 dB over the next best baseline on the Levin dataset at $\sigma = 0.1$). Without any training images, DAWN remains highly competitive with a state-of-the-art diffusion model on FFHQ, despite the latter being trained on a massive number of in-distribution images.

}


\date{\today}
\correspondence{\href{mailto:yidapan@umich.edu}{yidapan@umich.edu}}
\resources{\href{https://github.com/bamb00zled82/DAWN}{Code}}

\headerlogo{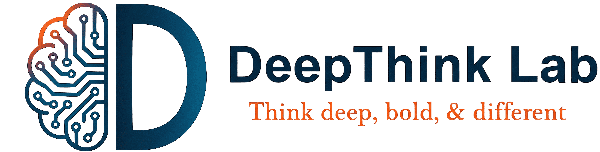}{https://deepthink-umich.github.io}

\begin{document}

\makeDeepthinkHeader

\begin{figure}[h]
\centering
\includegraphics[width=\linewidth]{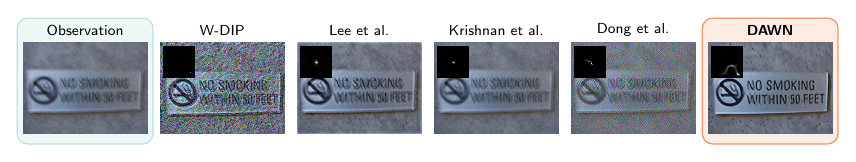}
\caption{Comparison on a blurry image from the real-world subset of \citet{lai2016comparative}, with noise of $\sigma = 0.1$. No ground truth image or kernel is available. Zoom in for best viewing. Additional non-cherry-picked real-world restored images are given in Figs.~\ref{fig:supp_real_flower}--\ref{fig:supp_real_text10} in the appendix.}
\label{fig:realworld}
\end{figure}

\newpage
\tableofcontents

\section{Introduction}
\label{sec:intro}

From consumer cameras to fluorescence and scanning electron microscopes, optical imaging systems routinely produce images degraded by spatially invariant blur~\citep{levin2009understanding, lee2024eccv}. In consumer photography, camera shake during long exposures introduces motion blur alongside significant photon and sensor noise~\citep{fergus2006removing, lai2016comparative}.
In scientific imaging, optical aberrations and diffraction-limited point spread functions (PSFs) produce blur that fundamentally limits the resolution of downstream biological or materials analysis~\citep{lee2024eccv, bredell2023wdip}.
In both settings, recovering the original image from a single degraded observation is essential, yet, in practice, the blur kernel is rarely known in advance~\citep{kundur1996blind}.
The degradation can be modeled as %
\begin{equation}
   \mathbf{y} = \mathbf{k} \ast \mathbf{x} + \mathbf{n},
   \label{eq:forward}
\end{equation}

\noindent where $\mathbf{x} \in \mathbb{R}^{n_x}$ is the image, $\mathbf{k} \in \mathbb{R}^{n_k}$ is the blur kernel (point spread function), $\ast$ denotes 2D convolution, and $\mathbf{n}$ is additive noise.
Recovering both $\mathbf{x}$ and $\mathbf{k}$ from the single degraded observation $\mathbf{y}$ is known as \emph{blind image deconvolution} (BID), a severely ill-posed inverse problem with broad practical importance~\citep{levin2009understanding, lai2016comparative}.

\begin{figure}[t]
\centering
\includegraphics[width=\linewidth]{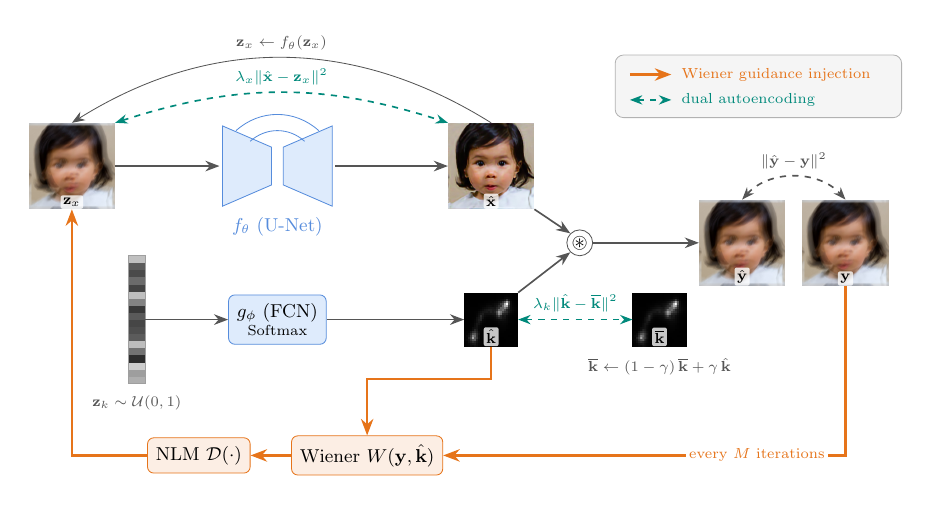}
\caption{Overview of the DAWN framework. The image and kernel networks are jointly optimized with a data fidelity loss and dual autoencoding regularization. Every $M$ outer iterations, the image network's input $\mathbf{z}_x$ is replaced with a Wiener-deconvolved and NLM-denoised estimate computed from the current kernel (direct injection). In between, the input is updated with the network's own output, so that the neural prior refines the injected estimate (neural refinement).}
\label{fig:architecture}
\end{figure}

BID becomes more challenging when the noise type is unknown and its levels are high. 
Under low or negligible noise, many existing approaches, from maximum a posteriori (MAP)-based optimization with handcrafted priors~\citep{fergus2006removing, krishnan2011blind, xu2013unnatural, pan2016blind} to learning-based methods~\citep{nah2017deep, zamir2021multi}, produce satisfactory results.
However, as the noise level increases, deblurring performance degrades substantially~\citep{liu2025iccv, lee2024eccv}.
This degradation represents a tradeoff. Focusing on removing the blur can severely corrupt the estimated image with amplified noise, whereas focusing on removing the noise tends to over-smooth the image and discard the fine details that accurate kernel estimation requires~\citep{lee2024eccv}.
In cases where the noise level cannot be accurately determined, which is typical in real-world problems, the task becomes more difficult.

Recently, the Deep Image Prior (DIP)~\citep{ulyanov2018dip} has emerged as an attractive paradigm for blind deblurring, as it requires no training data and instead leverages the implicit bias of untrained convolutional network architectures as an image prior, making it especially useful in settings where sufficiently large training data is lacking, such as in scientific imaging.
DIP-based blind deblurring methods such as SelfDeblur~\cite{ren2020selfdeblur} have demonstrated strong results under nearly noiseless settings by jointly optimizing image and kernel networks.
However, DIP is susceptible to noise overfitting~\citep{ulyanov2018dip, alkhouri2024aseqdip}, as the overparameterized network eventually fits the noise, causing the reconstruction quality to peak and then degrade~\citep{11480054}.
Several works have sought to improve the robustness of DIP-based methods, for example, by early stopping~\citep{huang2026principled} or by incorporating model-based guidance into the loss objective~\citep{bredell2023wdip, lee2024eccv}.

For BID, \citet{bredell2023wdip} proposed W-DIP, which adds a loss term that enforces consistency between the image network's output and a Wiener-deconvolved version of the observation, recomputed at every iteration. On the other hand, \citet{lee2024eccv} tackles a noisier BID problem setting, introducing an analytical %
kernel estimation function %
given a clear image estimate and the noisy observation. Consistency with these analytically derived kernels is enforced through one data-fidelity term per kernel at every iteration.
Both of these methods thus use model-based estimates only within the loss, which conveys their information to the network only as fast as gradient descent can fit it, a mechanism that underutilizes the information contained in these closed-form estimates.

Diffusion-based blind inverse solvers~\citep{chung2023blinddps, murata2023gibbsddrm} offer an alternative, as they learn the prior over clean training data. However, they face challenges of their own, including distribution shift under unfamiliar noise conditions, the need for large-scale clean training data, and a tendency to hallucinate fine details. These challenges are particularly problematic in scientific imaging applications, where reconstruction fidelity is critical~\citep{antun2020instabilities, bhadra2021hallucinations}.

In this work, we propose DAWN (\textbf{D}econvolution via \textbf{A}lternating \textbf{W}iener-guided \textbf{N}etworks), a training-data-free DIP-based blind deblurring framework that alternates between (i)~\emph{direct injection}, which replaces the network input with a denoised Wiener-deconvolved estimate computed from the current kernel, and (ii)~\emph{neural refinement}, which lets the network's own prior refine this estimate before the next injection.
During refinement, the image is free to depart from the Wiener estimate, which allows the measurements to correct the kernel (Sec.~\ref{sec:hard_injection}). Moreover, because DIP fits image structure before noise~\citep{11480054}, refinement removes much of the estimate's high-frequency noise and artifacts.
As the quality of the kernel directly influences the quality of the Wiener injection, we apply autoencoding regularization to both the image and kernel networks, enforcing self-consistency of each estimate throughout the joint optimization.
DAWN requires no knowledge of the noise level or type and consistently outperforms state-of-the-art methods on standard benchmarks under varying noise levels with one set of hyperparameters and no re-tuning. The main contributions are summarized next. 
\begin{enumerate}
  \item \textbf{DAWN: a training-data-free BID framework.} We propose DAWN, a training-data-free BID framework that alternates between directly injecting Wiener-deconvolved estimates into the input of an input-adaptive DIP and the neural refinement of these estimates. This uses model-based estimates more directly than loss-level guidance while avoiding the failure of injecting them at every iteration. To improve the reliability of these injections, we enforce autoencoding regularization of both image and kernel networks. %

  \item \textbf{Analysis of per-step Wiener injection.} We show that when the Wiener estimate is injected at every iteration, the image is tied to the current kernel and, in a simplified setting, the data-fidelity term itself favors the no-blur delta kernel regardless of the true blur, thus motivating our design choice of alternation with neural refinement. We also observe the resulting kernel collapse empirically and bound the deviation of the simplified data-fidelity term from the actual version.

  \item \textbf{Evaluations with dataless and data-dependent prior methods across various noise levels \& real-world images.} We provide comprehensive experiments on standard benchmarks under various levels of unknown Gaussian and Poisson noise, achieving significant improvement over the state of the art and very competitive results even under low noise, all without retuning. We also conduct a qualitative evaluation on real-world blurry images. 
\end{enumerate}

\section{Related Work}
\label{sec:related}

\subsection{MAP-Based Blind Deblurring}
\label{sec:related_map}

Classical approaches to blind deconvolution formulate the problem within a maximum a posteriori (MAP) framework, optimizing
\begin{equation}
  \min_{\mathbf{x}, \mathbf{k}} \; \|\mathbf{k} \ast \mathbf{x} - \mathbf{y}\|_2^2 + \lambda \mathcal{R}_x(\mathbf{x}) + \mu \mathcal{R}_k(\mathbf{k}),
  \label{eq:map}
\end{equation}
where $\mathcal{R}_x$ and $\mathcal{R}_k$ are image and kernel priors, respectively, weighted by the regularization parameters $\lambda > 0$ and $\mu > 0$, and the solution is typically obtained by alternating minimization with a coarse-to-fine strategy~\citep{cho2009fast, liu2025iccv}.
A variety of image priors have been proposed, including normalized sparsity~\citep{krishnan2011blind}, unnatural $\ell_0$ representations~\citep{xu2013unnatural}, dark channel priors~\citep{pan2016blind}, $\ell_0$-regularized intensity and gradient priors~\citep{pan2014l0}, total variation~\citep{perrone2014total}, and internal patch recurrence~\citep{michaeli2014blind}.
Edge-selection-based approaches~\citep{sun2013edge, cho2009fast} offer a complementary strategy by extracting salient edges to guide kernel estimation.
Kernel priors typically include non-negativity ($\mathbf{k} \geq 0$), unit-sum ($\sum_i k_i = 1$), and sparsity constraints.

While these methods perform well under noiseless settings, their reliance 
on image gradients and edge structures makes them fragile as noise 
increases~\citep{liu2025iccv}.
Several works specifically target the noisy setting. \citet{anger2019blind} adapt the $\ell_0$ framework to additive white Gaussian noise (AWGN) $\mathbf{n}$ in~\eqref{eq:forward} with standard deviation $\sigma = 0.1$, \citet{tai2012motion} iteratively alternate between motion-aware denoising and deblurring, and \citet{liu2025iccv} introduce a residual guidance strategy in the image pyramid.
However, such methods~\citep{anger2019blind, tai2012motion} 
require knowledge of the noise level, either as an explicit input 
or for internal parameter calibration, while others require careful per-dataset 
tuning~\citep{liu2025iccv}, and all remain limited by the 
expressiveness of their handcrafted priors.

\subsection{Data-Driven Approaches}
\label{sec:related_datadriven}

\paragraph{Supervised methods.}
End-to-end deep networks~\citep{nah2017deep, tao2018scale, kupyn2019deblurgan, zamir2021multi} have achieved strong performance by learning the mapping from blurry to sharp images on large paired datasets.
However, these models are constrained to the blur kernel distributions and noise levels seen during training, limiting their generalization to out-of-distribution degradations.

\paragraph{Diffusion-based methods.}
Diffusion posterior sampling (DPS)~\citep{chung2023dps} and its blind extension BlindDPS~\citep{chung2023blinddps} leverage pre-trained diffusion models as generative priors, while GibbsDDRM~\cite{murata2023gibbsddrm} employs a Laplace kernel prior that avoids assuming specific kernel distributions.
While these methods produce impressive perceptual quality, they face three challenges in the noisy blind deblurring setting.
First, they are susceptible to distribution shift when the noise level or type deviates significantly from the training conditions, which is a particular concern at the higher noise levels (e.g., $\sigma = 0.1$) that motivate this work.
Second, they require large-scale datasets of clean images for training, which are difficult to obtain in specialized scientific domains such as fluorescence microscopy~\citep{weigert2018content, krull2019noise2void}.
Third, they can hallucinate plausible but fictitious details~\citep{whang2022deblurdpm}, a property that is unacceptable in scientific imaging where reconstruction fidelity is essential.
These limitations motivate training-data-free approaches, which avoid the generalization limitations of learned priors.

\subsection{DIP-Based Blind Deblurring}
\label{sec:related_dip}

The Deep Image Prior (DIP)~\citep{ulyanov2018dip} demonstrates that the architecture of a convolutional neural network itself can serve as a powerful image prior, without any training data.
\citet{ren2020selfdeblur} extended this idea to blind deblurring in SelfDeblur, which uses a U-Net to generate the image estimate $f_\theta(\mathbf{z}_x)$ and a fully connected network (FCN) to generate the kernel estimate $g_\phi(\mathbf{z}_k)$, optimized jointly as
\begin{equation}
  \min_{\theta, \phi} \; \|f_\theta(\mathbf{z}_x) \ast g_\phi(\mathbf{z}_k) - \mathbf{y}\|_2^2,
  \label{eq:selfdeblur}
\end{equation}
where $\mathbf{z}_x$ and $\mathbf{z}_k$ are fixed random input tensors.
Despite strong results under low noise, DIP-based methods are prone to noise overfitting~\citep{ulyanov2018dip, alkhouri2024aseqdip}.

Several methods have sought to improve the robustness and performance of DIP-based methods in BID. 
\citet{kotera2021improving} initialize SelfDeblur from a coarser-scale solution and add conventional regularization terms.
\citet{bredell2023wdip} proposed W-DIP, which augments~\eqref{eq:selfdeblur} with auxiliary terms $\alpha\|f_\theta(\mathbf{z}_x) - W(\mathbf{y}, \mathbf{k}_{\mathrm{aux}})\|_2^2 + \beta\|g_\phi(\mathbf{z}_k) - \mathbf{k}_{\mathrm{aux}}\|_2^2$, where $W(\mathbf{y}, \mathbf{k})$ denotes Wiener deconvolution and $\mathbf{k}_{\mathrm{aux}}$ is an auxiliary kernel introduced via half-quadratic splitting.
\citet{lee2024eccv} instead introduce a noise-robust kernel estimation function that analytically computes the kernel given a clear image estimate and a noise suppression factor, using DIP to provide the clear image, and achieve substantially improved performance under nontrivial noise.
In the latter two methods, the model-based estimate enters only through the loss. DAWN instead periodically sets the network input to the estimate, which then also serves as the image autoencoding target, and alternates this direct injection with neural refinement (Secs.~\ref{sec:hard_injection} and~\ref{sec:dawn_framework}).

\section{Proposed Method}
\label{sec:method}

\subsection{Preliminaries: Input-Adaptive DIP}
\label{sec:prelim}
Standard DIP methods~\citep{ulyanov2018dip, ren2020selfdeblur} parameterize the image estimate as $\hat{\mathbf{x}} = f_\theta(\mathbf{z})$, where $\mathbf{z} \sim \mathcal{U}(0, 1)$ is a random input tensor that remains fixed during the optimization.
The network learns a mapping from noise space to image space, so the input $\mathbf{z}$ carries no image-level structure.

The Autoencoding Sequential DIP (aSeqDIP) 
framework~\citep{alkhouri2024aseqdip} takes a fundamentally 
different approach. Originally developed for non-blind inverse 
problems, where the goal is to recover $\mathbf{x}$ from 
$\mathbf{y} \approx \mathbf{A}\mathbf{x}$ given a known forward 
operator $\mathbf{A}$, aSeqDIP initializes the network input 
with %
$\mathbf{z}^{(0)} = \mathbf{A}^H \mathbf{y}$, and sequentially updates it across $K$ outer 
iterations.
At each outer iteration $k = 1,\dots,K$, the network parameters $\theta$ are updated for $N$ gradient steps on the objective
\begin{equation}
  \min_{\theta} \|\mathbf{A} f_\theta(\mathbf{z}^{(k-1)}) - \mathbf{y}\|_2^2 + \lambda \|f_\theta(\mathbf{z}^{(k-1)}) - \mathbf{z}^{(k-1)}\|_2^2,
  \label{eq:aseqdip}
\end{equation}
where the first term enforces data consistency and the second is an \emph{autoencoding} regularizer that encourages the network output to remain consistent with its input, mitigating noise overfitting.
After optimization, the input is updated as $\mathbf{z}^{(k)} = f_\theta(\mathbf{z}^{(k-1)})$, so that subsequent iterations operate on progressively refined estimates.
Critically, the input $\mathbf{z}^{(k)}$ is always a structured image estimate rather than a fixed random tensor, which DAWN exploits to inject model-based estimates directly.

\subsection{Where and When to Inject Wiener Guidance}
\label{sec:hard_injection}
\begin{figure}[t]
\centering
\includegraphics[width=\textwidth]{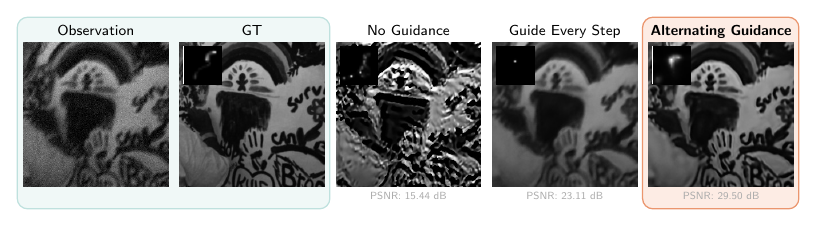}
\caption{Direct injection on a Levin image ($\sigma = 0.05$). Without guidance, the reconstruction is dominated by artifacts. Injecting the Wiener estimate at every step collapses the kernel to a single pixel, and the output remains a blurry copy of the observation. This motivates our design choice of alternating direct injection with neural refinement, which circumvents this collapse.}
\label{fig:hard_injection}
\end{figure}

\paragraph{Where to inject.}
Existing DIP-based methods use a model-based estimate only within the loss, such as the Wiener-deconvolved image in W-DIP~\citep{bredell2023wdip} and the analytically computed kernel in \citet{lee2024eccv}. 
In this work, we propose directly injecting the model-based estimate into the network input to replace it, which input-adaptive DIP (Sec.~\ref{sec:prelim}) makes possible.

For blind deblurring with image network $f_\theta$, image input $\mathbf{z}_x$, and a jointly estimated kernel $\hat{\mathbf{k}}$ (Sec.~\ref{sec:dawn_framework}), a natural estimate is the Wiener-deconvolved observation computed with the current kernel,
\begin{equation}
  W(\mathbf{y}, \hat{\mathbf{k}}) = \mathcal{F}^{-1}\!\left\{
  \frac{\overline{\hat{K}}\, Y}{|\hat{K}|^2 + C}
  \right\},
  \label{eq:wiener}
\end{equation}
where capitals denote Fourier transforms ($\hat{K} = \mathcal{F}\{\hat{\mathbf{k}}\}$, $Y = \mathcal{F}\{\mathbf{y}\}$), $\overline{\,\cdot\,}$ denotes complex conjugation, and $C > 0$ is a regularization constant.
\emph{Direct injection} replaces the network input with a denoised version of this estimate, $\mathbf{z}_x \leftarrow \mathcal{D}(W(\mathbf{y}, \hat{\mathbf{k}}))$, where $\mathcal{D}$ is a non-local means (NLM) denoiser~\citep{zhong2013cvpr}. Here we note that the addition of $\mathcal{D}$ is not critical to DAWN's strong performance; without it, DAWN still outperforms all baselines. See Table~\ref{tab:abl_nlm}. 
Unlike an auxiliary loss term, which enforces the network output toward the estimate only gradually through gradient updates, direct injection makes the estimate the starting point of the subsequent optimization. 

\paragraph{Challenges of per-step injection.}
The most straightforward design choice would be to apply this model-based guidance at every step, as previous works do through the loss~\citep{bredell2023wdip, lee2024eccv}. For direct injection, however, we argue below that doing so biases the kernel toward a Dirac delta, which is the no-blur solution that sparse image priors are known to favor in joint image--kernel estimation~\citep{levin2009understanding}. To illustrate this, we consider a simplified setting where the image estimate equals the injected Wiener estimate at each step, $\hat{\mathbf x} = W(\mathbf y,\hat{\mathbf k})$, and ignore the NLM denoiser (Appendix~\ref{sec:supp_residual_scope} bounds the error of this simplification). Substituting~\eqref{eq:wiener}, the data-fidelity term becomes
\begin{equation}
  \|\hat{\mathbf{k}} \ast W(\mathbf{y}, \hat{\mathbf{k}}) - \mathbf{y}\|_2
  = \Big\| \mathcal{F}^{-1}\Big\{ \frac{C}{|\hat{K}|^2 + C}\, Y \Big\} \Big\|_2
  \ge \frac{C}{1 + C}\, \|\mathbf{y}\|_2
  = \|\boldsymbol{\delta} \ast W(\mathbf{y}, \boldsymbol{\delta}) - \mathbf{y}\|_2,
  \label{eq:self_residual}
\end{equation}
where $\boldsymbol{\delta}$ is the Dirac delta kernel and here $\ast$ is circular convolution. The bound follows from Parseval's theorem and $|\hat{K}(\omega)| \le \sum_j \hat{k}_j = 1$, which holds for any nonnegative kernel that sums to one. 
Because the injection ties the image to $\hat{\mathbf k}$, the data-fidelity term itself favors the delta even over the true kernel, and thus biases $\hat{\mathbf k}$ toward the no-blur solution. Empirically, we also observe this behavior on Levin, where per-step injection indeed collapses all kernels to a near-delta (Fig.~\ref{fig:hard_injection} and Table~\ref{tab:kernel_collapse}), and PSNR drops from 25.02 to 22.59\,dB, erasing most of DAWN's advantage over the next best baseline (Table~\ref{tab:abl_lx_M}, right).

\paragraph{When to inject.}
The intuition established above suggests avoiding per-step injection. In this work, we thus propose alternating direct injection with refinement by the neural prior.
Between injections, the input evolves under the network ($\mathbf{z}_x \leftarrow f_\theta(\mathbf{z}_x)$), so the image is free to depart from the Wiener estimate and the data-fidelity term can again constrain the kernel.
Moreover, since DIP fits image structure before noise~\citep{ulyanov2018dip}, these updates suppress much of the high-frequency noise and artifacts of the injected estimate. The sensitivity of Wiener deconvolution to observation noise and kernel estimation errors is bounded in Appendix~\ref{sec:supp_wiener_sensitivity}. 

\subsection{DAWN: Deconvolution via Alternating Wiener Guidance}
\label{sec:dawn_framework}

\paragraph{Overall objective.}
DAWN jointly optimizes the parameters of an image network $f_\theta$ (U-Net) and a kernel network $g_\phi$ (FCN with softmax output, ensuring the physical constraints $\sum k_i = 1$ and $k_i \geq 0$ that any valid point spread function must satisfy).
At each gradient step within outer iteration $k$, the parameters $(\theta, \phi)$ are updated to minimize
\begin{equation}
  \min_{\theta, \phi} \;\; {\underbrace{\|f_\theta(\mathbf{z}_x) \ast g_\phi(\mathbf{z}_k) - \mathbf{y}\|_2^2}_{\text{data fidelity}}}
  + {\underbrace{\lambda_x \|f_\theta(\mathbf{z}_x) - \mathbf{z}_x\|_2^2}_{\text{image autoencoding}}}
  + {\underbrace{\lambda_k \|g_\phi(\mathbf{z}_k) - \overline{\mathbf{k}}\|_2^2}_{\text{kernel autoencoding}}},
  \label{eq:dawn_loss}
\end{equation}
where $\overline{\mathbf{k}}$ is an exponential moving average of the kernel estimates (defined in Section~\ref{sec:kernel_ae}) and $\mathbf{z}_k$ is a fixed random input to the kernel network.
We denote the image and kernel outputs as $\hat{\mathbf{x}} = f_\theta(\mathbf{z}_x)$ and $\hat{\mathbf{k}} = g_\phi(\mathbf{z}_k)$, respectively.
Unlike in prior DIP methods, the image input $\mathbf{z}_x$ is \emph{not} a fixed random tensor, and DAWN exploits it for the direct injection of model-based estimates, as described next and illustrated in Fig.~\ref{fig:architecture}.

\paragraph{Alternating direct injection and neural refinement.}
Motivated by Section~\ref{sec:hard_injection}, DAWN alternates between two phases. Every $M$ outer iterations, a \emph{direct injection} replaces the network input with the denoised Wiener estimate computed from the current kernel, and in all other outer iterations, \emph{neural refinement} updates the input with the network's own output:
\begin{equation}
  \mathbf{z}_x^{(k)} =
  \begin{cases}
    \mathcal{D}\!\left(W(\mathbf{y},\, \hat{\mathbf{k}}^{(k)})\right) & \text{if } k \bmod M = 0, \\[4pt]
    f_\theta(\mathbf{z}_x^{(k-1)}) & \text{otherwise},
  \end{cases}
  \label{eq:zx_update}
\end{equation}
with $W$ from~\eqref{eq:wiener} and $\mathcal{D}$ the NLM denoiser.
Right after an injection, the image autoencoding term in~\eqref{eq:dawn_loss} keeps the network output close to the injected estimate. During refinement, the input, and with it the autoencoding target, follows the network's output. As discussed in Section~\ref{sec:hard_injection}, this enables the measurements to correct the kernel again and lets the neural prior remove much of the estimate's high-frequency noise and artifacts.
The period $M$ balances the two phases. Setting $M = 1$ reduces DAWN to direct injection at every iteration (Section~\ref{sec:hard_injection}), whereas a very large $M$ supplies too little guidance within the optimization budget (see the ablation over $M$ in Table~\ref{tab:abl_lx_M}, right).

\subsection{Kernel Autoencoding}
\label{sec:kernel_ae}

The effectiveness of the Wiener injection in~\eqref{eq:zx_update} depends critically on the quality of the kernel estimate $\hat{\mathbf{k}}^{(k)}$, as a Wiener-deconvolved image computed with a poor kernel can be actively harmful.
We therefore additionally regularize the kernel network toward its own recent history.
Specifically, we maintain an exponential moving average (EMA) of the kernel estimates:
\begin{equation}
  \overline{\mathbf{k}}^{(k)} = (1 - \gamma) \cdot \overline{\mathbf{k}}^{(k-1)} + \gamma \cdot \hat{\mathbf{k}}^{(k)},
  \label{eq:ema}
\end{equation}
where $\gamma \in (0, 1)$ is a momentum parameter controlling the update rate and $\overline{\mathbf{k}}^{(0)}$ is initialized with the first kernel output.
The kernel autoencoding term $\lambda_k\|g_\phi(\mathbf{z}_k) - \overline{\mathbf{k}}\|_2^2$ penalizes the kernel network for deviating from its own running average, enforcing self-consistency without anchoring to any external target. Table~\ref{tab:ablation} ablates this EMA-based term.
Our complete algorithm is given in Algorithm~\ref{alg:dawn}.

\begin{algorithm}[t]
\caption{DAWN}
\label{alg:dawn}
\begin{algorithmic}[1]
\REQUIRE $\mathbf{y}$, $K$, $N$, $M$, $\lambda_x$, $\lambda_k$, $\gamma$
\STATE Initialize $f_\theta$, $g_\phi$;\; 
$\mathbf{z}_x \leftarrow \mathbf{y}$;\; 
$\mathbf{z}_k \sim \mathcal{U}(0, 1)$; $\overline{\mathbf{k}} \leftarrow g_\phi(\mathbf{z}_k)$
\FOR{$k = 1, \ldots, K$}
  \FOR{$n = 1, \ldots, N$}
    \STATE $\hat{\mathbf{x}} \leftarrow f_\theta(\mathbf{z}_x + \epsilon)$, $\epsilon \sim \mathcal{N}(0, \sigma_z^2 \mathbf{I})$; \; $\hat{\mathbf{k}} \leftarrow g_\phi(\mathbf{z}_k)$
    \STATE $\mathcal{L} \leftarrow \|\hat{\mathbf{x}} \ast \hat{\mathbf{k}} - \mathbf{y}\|^2 + \lambda_x\|\hat{\mathbf{x}} - \mathbf{z}_x\|^2 + \lambda_k\|\hat{\mathbf{k}} - \overline{\mathbf{k}}\|^2$
    \STATE Update $(\theta, \phi)$ via Adam on $\mathcal{L}$
  \ENDFOR
  \STATE $\overline{\mathbf{k}} \leftarrow (1 - \gamma)\,\overline{\mathbf{k}} + \gamma\,\hat{\mathbf{k}}$
  \DAWNcomment{Kernel EMA}
  \IF{$k \bmod M = 0$}
    \STATE $\mathbf{z}_x \leftarrow \mathcal{D}(W(\mathbf{y},\, \hat{\mathbf{k}}))$
    \DAWNcomment{Wiener injection}
  \ELSE
    \STATE $\mathbf{z}_x \leftarrow f_\theta(\mathbf{z}_x)$
    \DAWNcomment{Neural refinement}
  \ENDIF
\ENDFOR
\RETURN $\hat{\mathbf{x}} = f_\theta(\mathbf{z}_x)$, \; $\hat{\mathbf{k}} = g_\phi(\mathbf{z}_k)$
\end{algorithmic}
\end{algorithm}

\section{Experiments}
\label{sec:experiments}

\begin{figure}[t]
\centering
\includegraphics[width=\textwidth]{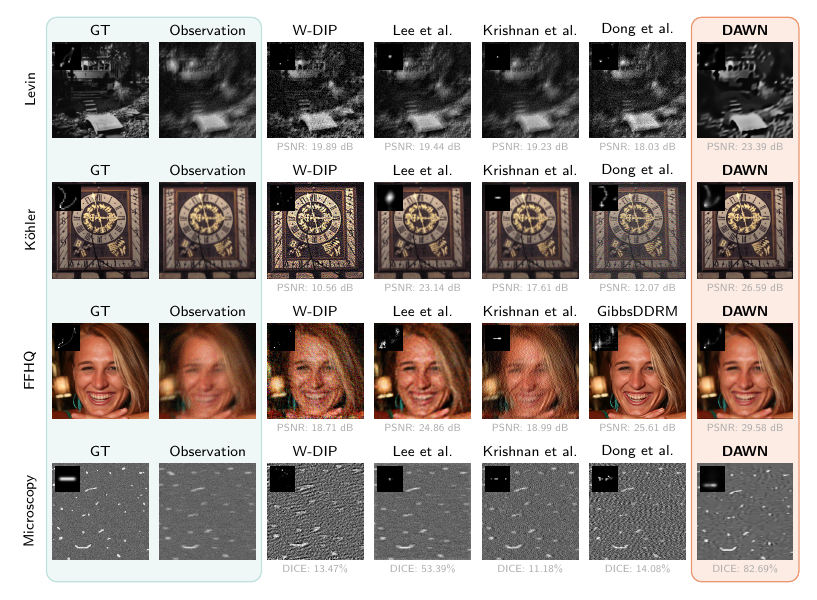}
\caption{BID results 
across four datasets at $\sigma = 0.05$ Gaussian noise. DAWN produces 
sharper reconstructions with more accurate kernel estimates 
across all datasets. For the FFHQ row, Dong et al.\ is 
replaced with GibbsDDRM~\cite{murata2023gibbsddrm}, whose 
diffusion prior is pre-trained on FFHQ. The microscopy row 
reports DICE (segmentation fidelity) in place of PSNR. Zoom in for best viewing.}
\label{fig:qualitative}
\end{figure}

We evaluate DAWN across a diverse set of blind deblurring scenarios under a range of nontrivial noise levels.
On the Levin~\citep{levin2009understanding} and K\"ohler~\citep{kohler2012recording} benchmarks, we test under AWGN at two levels ($\sigma = 0.05$ and $\sigma = 0.1$), and on Levin we additionally test four noise levels.
On a microscopy dataset with synthetic blur~\citep{bredell2023wdip}, we evaluate deblurring quality through downstream segmentation fidelity, and on the FFHQ dataset~\citep{karras2019ffhq} we compare against a diffusion-based approach, and we also include real-world blurry images~\citep{lai2016comparative}.
Fig.~\ref{fig:qualitative} presents qualitative comparisons across all datasets.
We evaluate using PSNR, SSIM, and maximum normalized correlation (MNC) for kernel accuracy, and for microscopy we additionally report DICE for segmentation fidelity. Formal definitions of MNC and DICE are provided in Appendix~\ref{sec:supp_metrics}.

\paragraph{Implementation details.}
DAWN uses the same U-Net image generator and FCN kernel generator as SelfDeblur~\cite{ren2020selfdeblur}.
We run $K = 2000$ outer iterations of $N = 2$ gradient steps (4000 steps in total), with injection period $M = 200$ and image autoencoding weight $\lambda_x = 1$ chosen by the ablations in Table~\ref{tab:abl_lx_M}.
Across all experiments and all noise levels, DAWN uses a single set of hyperparameters, listed in Appendix~\ref{sec:supp_implementation}, and requires no knowledge of the noise level or type.
DAWN also uses no early stopping and we always report its final iterate, which remains stable (Fig.~\ref{fig:convergence}).

\paragraph{Baselines.}
We compare against the DIP-based SelfDeblur~\cite{ren2020selfdeblur}, W-DIP~\cite{bredell2023wdip}, and \citet{lee2024eccv}, and the classical MAP methods of \citet{krishnan2011blind}, Pan-DCP~\cite{pan2016blind}, and \citet{dong2017outlier}. On FFHQ, we also compare against the diffusion-based GibbsDDRM~\cite{murata2023gibbsddrm}.
All baselines use their official implementations with default hyperparameters, without tuning for the noise level (GibbsDDRM, which requires the noise level as input, is given the true value), and all methods are given the size of the ground-truth kernel.
Like DAWN, every method is evaluated at its final output, without oracle early stopping.
We exclude \citet{liu2025iccv}, as no public implementation is available and the method requires per-dataset hyperparameter tuning, as noted by the authors. We exclude \citet{zhuang2024ijcv} for the same no-public-code reason.
Detailed baseline descriptions are provided in Appendix~\ref{sec:supp_baselines}.

\subsection{Results}
\label{sec:results}

We conduct extensive experiments on standard benchmarks, primarily under Gaussian noise. DAWN also outperforms all baselines under Poisson shot noise (Fig.~\ref{fig:shot_noise}), and the appendix further provides runtimes (Table~\ref{tab:runtime}), ablations (Appendix~\ref{sec:supp_ablation}), and per-dataset visualizations (Figs.~\ref{fig:supp_real_flower}--\ref{fig:qual_micro}).

\begin{table}[t]
\centering
\caption{Quantitative comparison of blind deblurring results under Gaussian noise on standard benchmarks: Levin~\citep{levin2009understanding} and K\"ohler~\citep{kohler2012recording} datasets. \textbf{Bold}: best.}
\label{tab:gaussian}
\footnotesize
\setlength{\tabcolsep}{3pt}
\begin{tabular}{l | cccccc | cccccc}
\toprule
 & \multicolumn{6}{c|}{\textbf{Levin}} & \multicolumn{6}{c}{\textbf{K\"ohler}} \\
 & \multicolumn{3}{c}{$\sigma = 0.05$} & \multicolumn{3}{c|}{$\sigma = 0.1$} & \multicolumn{3}{c}{$\sigma = 0.05$} & \multicolumn{3}{c}{$\sigma = 0.1$} \\
Method & PSNR$\uparrow$ & SSIM$\uparrow$ & MNC$\uparrow$ & PSNR$\uparrow$ & SSIM$\uparrow$ & MNC$\uparrow$ & PSNR$\uparrow$ & SSIM$\uparrow$ & MNC$\uparrow$ & PSNR$\uparrow$ & SSIM$\uparrow$ & MNC$\uparrow$ \\
\midrule
\rowcolor{blue!8}
\best{DAWN (Ours)  }     & \best{25.02} & \best{0.75} & \best{0.70} & \best{23.59} & \best{0.65} & \best{0.65} & \best{26.48} & \best{0.74} & \best{0.62} & \best{25.43} & \best{0.72} & \best{0.61} \\
SelfDeblur & 20.85 & 0.42 & 0.47 & 18.88 & 0.29 & 0.46 & 16.95 & 0.23 & 0.48 & 12.55 & 0.11 & 0.45 \\
W-DIP         & 21.02 & 0.42 & 0.49 & 19.07 & 0.30 & 0.47 & 16.67 & 0.22 & 0.49 & 12.21 & 0.10 & 0.47 \\
Lee et al.        & 21.96 & 0.53 & 0.56 & 19.09 & 0.32 & 0.52 & 24.87 & 0.63 & 0.51 & 24.34 & 0.62 & 0.49 \\
Krishnan et al.    & 20.01 & 0.35 & 0.50 & 16.72 & 0.19 & 0.53 & 21.69 & 0.31 & 0.45 & 17.79 & 0.14 & 0.45 \\
Pan-DCP          & 11.97 & 0.14 & 0.57 & 9.48 & 0.06 & 0.55 & 12.86 & 0.09 & 0.53 & 10.21 & 0.04 & 0.51 \\
Dong et al.   & 16.06 & 0.22 & 0.61 & 13.91 & 0.14 & 0.54 & 16.05 & 0.13 & 0.57 & 13.36 & 0.07 & 0.50 \\
\bottomrule
\end{tabular}
\end{table}

\paragraph{Levin.}
The Levin dataset consists of 4 grayscale images convolved with 8 blur kernels ($13 \times 13$ to $27 \times 27$), yielding 32 pairs. Table~\ref{tab:gaussian} (left) shows DAWN leads on all metrics, surpassing the next best by 4.5 dB at $\sigma = 0.1$. Without Wiener injection, PSNR at $\sigma = 0.05$ drops from 25.02 to 21.15\,dB (Table~\ref{tab:ablation}).

\begin{figure}[t]
\centering
\includegraphics[width=\linewidth]{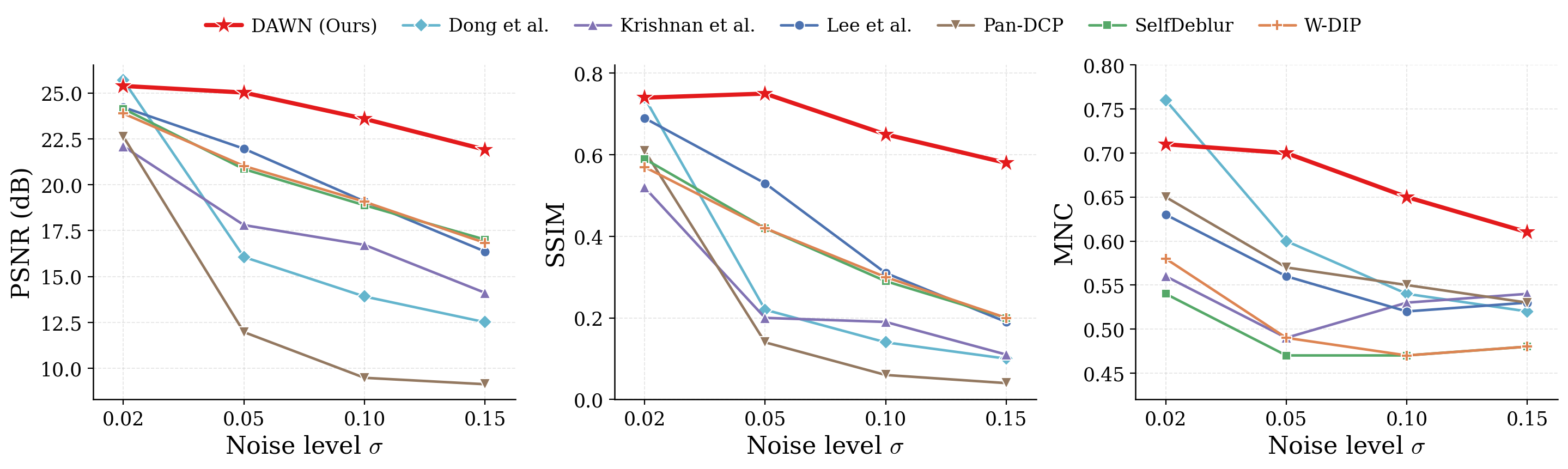}
\caption{Performance across noise levels on Levin~\citep{levin2009understanding}. DAWN maintains the best metrics for $\sigma \geq 0.05$ without requiring knowledge of the noise level.}
\label{fig:noise_robustness}
\end{figure}

\paragraph{Robustness across noise levels.}
Fig.~\ref{fig:noise_robustness} extends the Gaussian noise comparison on Levin to $\sigma \in \{0.02, 0.05, 0.10, 0.15\}$. Classical methods such as \citet{dong2017outlier} are competitive only at $\sigma = 0.02$ and collapse rapidly, whereas DAWN degrades gracefully, maintains the best PSNR, SSIM, and MNC for $\sigma \geq 0.05$, and remains highly competitive at $\sigma = 0.02$ despite being tuned for moderate-to-heavy noise.

\paragraph{K\"ohler.}
This benchmark contains 4 scenes and 12 real camera-shake trajectories (48 pairs), with kernels ranging from under $10 \times 10$ to over $100 \times 100$, and both extremes are challenging.
We convolve each kernel with the sharp image following \citet{deblursdi} and add Gaussian noise.
Table~\ref{tab:gaussian} (right) shows that DAWN achieves the best performance on all metrics, and the gap over SelfDeblur and W-DIP widens considerably in this extreme-kernel regime.

\paragraph{Microscopy.}
On the dataset from~\citep{bredell2023wdip} (real microscopy images with synthetic kernels) at $\sigma = 0.05$, Table~\ref{tab:micro_ffhq} shows DAWN achieves the highest DICE, PSNR, and MNC.

\begin{table}[t]
\centering
\caption{Results on FFHQ~\citep{karras2019ffhq} and Microscopy~\citep{bredell2023wdip}. GibbsDDRM uses an FFHQ pre-trained prior and requires knowledge of noise level. \textbf{Bold}: best.}
\label{tab:micro_ffhq}
\setlength{\tabcolsep}{5pt}
\small
\begin{tabular}{l | ccc | ccc}
\toprule
 & \multicolumn{3}{c|}{\textbf{FFHQ}} & \multicolumn{3}{c}{\textbf{Microscopy}} \\
Method & PSNR$\uparrow$ & SSIM$\uparrow$ & MNC$\uparrow$ & DICE$\uparrow$ & PSNR$\uparrow$ & MNC$\uparrow$ \\
\midrule
\rowcolor{blue!8}
\best{DAWN (Ours)} & \best{27.95} & 0.81 & \best{0.75} & \best{72.97\%} & \best{19.69} & \best{0.84} \\
GibbsDDRM & 27.47 & \best{0.83} & 0.73 & --- & --- & --- \\
Lee et al. & 25.93 & 0.64 & 0.73 & 58.72\% & 19.32 & 0.39 \\
SelfDeblur & 21.50 & 0.37 & 0.49 & 41.20\% & 16.48 & 0.30 \\
W-DIP & 21.15 & 0.35 & 0.49 & 43.16\% & 17.61 & 0.30 \\
Krishnan et al. & 21.25 & 0.33 & 0.47 & 34.42\% & 17.23 & 0.34 \\
Pan-DCP & 11.74 & 0.09 & 0.55 & 18.53\% & 11.56 & 0.64 \\
Dong et al. & 15.52 & 0.15 & 0.64 & 24.88\% & 15.31 & 0.49 \\
\bottomrule
\end{tabular}
\end{table}

\begin{figure}[t]
\centering
\includegraphics[width=0.5\linewidth]{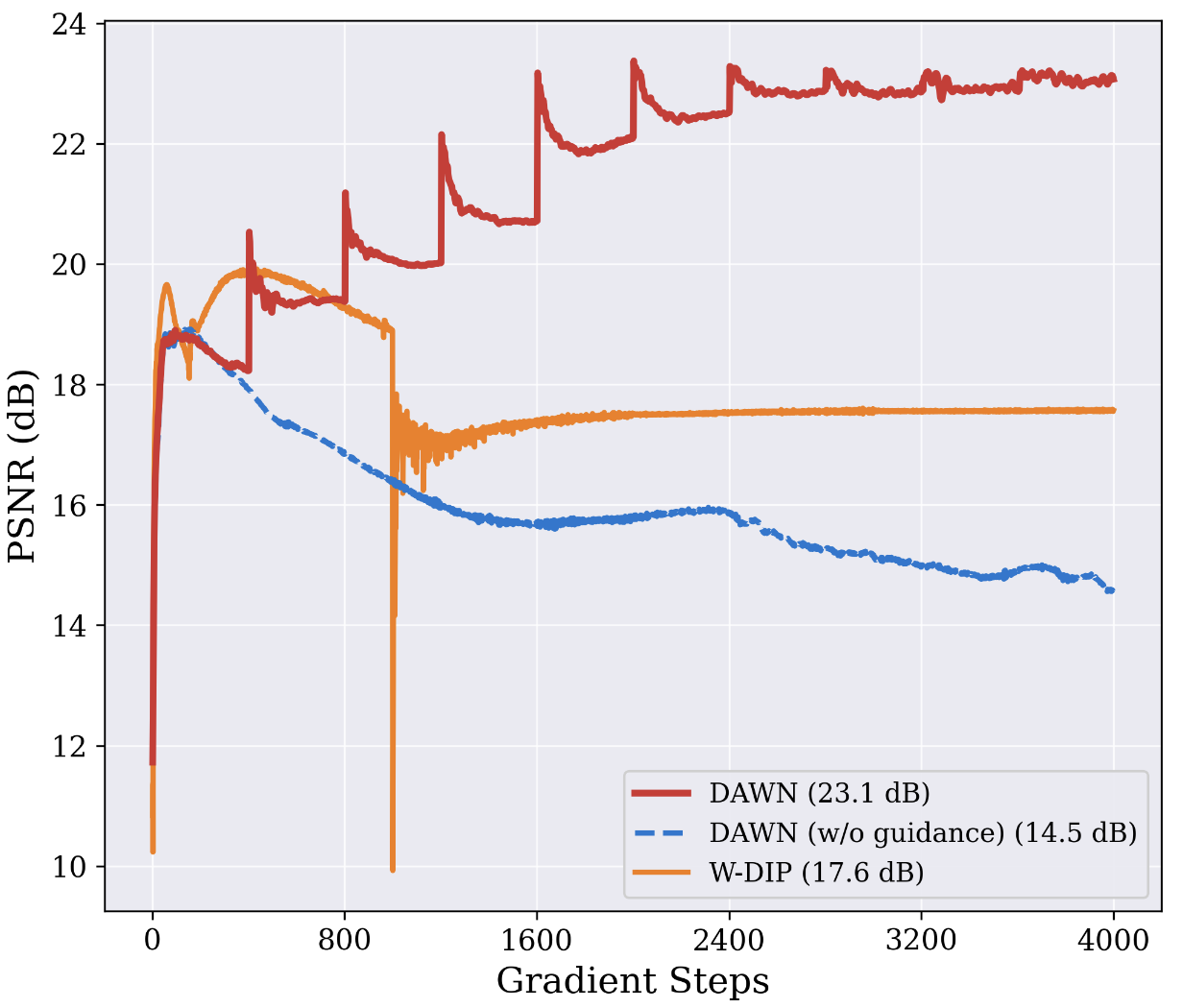}
\caption{Convergence of a Levin image ($\sigma = 0.05$). DAWN's Wiener injections appear as vertical jumps in the red curve. W-DIP switches from an MSE to an SSIM loss at iteration 1000.}
\label{fig:convergence}
\end{figure}

\paragraph{FFHQ.}
Alongside existing baselines, we benchmark against GibbsDDRM~\cite{murata2023gibbsddrm} on 80 images (first 10 FFHQ faces $\times$ 8 realistic motion blur kernels from Levin, $\sigma = 0.05$), a setting that is fully in-distribution for GibbsDDRM, whose image prior is pretrained on FFHQ and Laplace kernel prior accommodates arbitrary kernel shapes. DAWN outperforms GibbsDDRM by $\sim$0.5 dB without any training data or knowledge of the noise level, both of which GibbsDDRM uses.

\paragraph{Real-world.}
Fig.~\ref{fig:realworld} shows results on a real blurry image that is not guaranteed to have spatially invariant blur or to follow the simplified convolution model in~\eqref{eq:forward}. As the kernel size is unknown, we follow W-DIP~\cite{bredell2023wdip} and set the kernel support to $51 \times 51$ for all methods. DAWN recovers visibly sharper image and kernel structures with minimal artifacts. Figs.~\ref{fig:supp_real_flower}--\ref{fig:supp_real_text10} in the appendix show more real-world results, where DAWN clearly outperforms the baselines with better kernel estimates and clearer images.

\paragraph{Convergence.}
Fig.~\ref{fig:convergence} tracks PSNR over the optimization on a Levin image. DAWN improves with every injection and ends close to its best iterate, whereas DAWN without guidance and W-DIP peak early and then deteriorate. Although W-DIP also uses Wiener deconvolution as guidance, DAWN exploits this guidance more effectively and substantially outperforms W-DIP.

\section{Conclusion}
\label{sec:conclusion}

We presented DAWN, a training-data-free blind deblurring framework that alternates between direct injection of Wiener-deconvolved estimates into the network input and neural refinement of these estimates by the network's own prior.
Experiments show that DAWN consistently outperforms state-of-the-art methods under Gaussian and Poisson noise without knowledge of the noise level or retuning, and that it remains stable without early stopping.
DAWN currently inherits the per-image computational cost of DIP-based optimization. Future work includes improving speed and performance in the low-noise regime.

\section*{Acknowledgements}

YP acknowledges support from the Research Experiences for Undergraduates (REU) program under NSF Grant No. CCF-2621730. QQ and YJ acknowledge support from the NSF under Grant Nos. 2212066 and 2143904. IA acknowledges support from the U.S. Department of Energy through the Laboratory Directed Research and Development at Los Alamos National Laboratory (LANL) under project 20250048DR. LANL is operated by Triad National Security, LLC, for the National Nuclear Security Administration of U.S. Department of Energy (Contract No. 89233218CNA000001).

\printbibliography

\newpage
\appendix
\crefalias{section}{appendix}
\crefalias{subsection}{appendix}

\section{Analysis of Wiener Guidance}
\label{sec:supp_theory}
In Sec.~\ref{sec:hard_injection}, we analyze a simplified Wiener-based re-blurring model, in which the data-fidelity term favors the no-blur solution under per-step injection. Here, we complement this argument in two ways. Sec.~\ref{sec:supp_wiener_sensitivity} analyzes the sensitivity of Wiener guidance, and Sec.~\ref{sec:supp_residual_scope} bounds the discrepancy between the simplified residual and that of an actual image estimate, which clarifies the role of neural refinement and the scope of the simplified argument.

\subsection{Sensitivity of Wiener Guidance}
\label{sec:supp_wiener_sensitivity}

For a fixed regularization constant $C>0$, we consider the Wiener operator
\begin{equation}
W(\mathbf{y},\mathbf{k})
=\mathcal{F}^{-1}\!\left\{
\frac{\overline{K}\,Y}{|K|^2+C}
\right\},
\label{eq:supp_wiener}
\end{equation}
where $K=\mathcal{F}\{\mathbf{k}\}$ and $Y=\mathcal{F}\{\mathbf{y}\}$, with the kernel zero-padded to the image grid. We use the unnormalized forward discrete Fourier transform and its normalized inverse.

The following perturbation estimate is a straightforward consequence of standard spectral-filter bounds for Tikhonov regularization, Parseval's identity, and the Fourier convolution theorem (see \cite[Chap.~4]{hansen2010discrete} and \cite[Chap.~5]{vogel2002computational}). We state it explicitly to quantify the dependence of DAWN's Wiener guidance on observation perturbations and kernel-estimation errors.

\begin{proposition}[Sensitivity to observation and kernel perturbations]
\label{prop:supp_wiener_sensitivity}
For any observation perturbation $\mathbf{e}$ and kernels $\mathbf{k}_1,\mathbf{k}_2$ on the same grid, we have
\begin{equation}
\left\|W(\mathbf{y}+\mathbf{e},\mathbf{k}_2)
-W(\mathbf{y},\mathbf{k}_1)\right\|_2
\leq
\frac{\|\mathbf{e}\|_2}{2\sqrt{C}}
+\frac{\|\mathbf{y}\|_2}{C}
\|\mathbf{k}_2-\mathbf{k}_1\|_1.
\label{eq:supp_wiener_sensitivity}
\end{equation}
\end{proposition}

\begin{proof}
For a fixed kernel, the operator is linear in the observation. Its largest Fourier gain satisfies
\begin{equation}
\sup_{r\geq0}\frac{r}{r^2+C}=\frac{1}{2\sqrt{C}}.
\label{eq:supp_wiener_gain}
\end{equation}
Parseval's identity therefore bounds the effect of changing the observation by $\|\mathbf{e}\|_2/(2\sqrt{C})$.

For the kernel perturbation, define $h(a)=\overline{a}/(|a|^2+C)$ for a complex number $a$. Direct subtraction gives
\begin{equation}
h(a)-h(b)
=\frac{C(\overline{a}-\overline{b})
+\overline{a}\,\overline{b}(b-a)}
{(|a|^2+C)(|b|^2+C)}.
\label{eq:supp_scalar_difference}
\end{equation}
Consequently,
\begin{equation}
|h(a)-h(b)|
\leq
\frac{C+|a||b|}{(|a|^2+C)(|b|^2+C)}|a-b|
\leq\frac{1}{C}|a-b|.
\label{eq:supp_scalar_lipschitz}
\end{equation}
The second inequality follows from
\begin{equation*}
(|a|^2+C)(|b|^2+C)-C(C+|a||b|)
=|a|^2|b|^2+C(|a|^2+|b|^2-|a||b|)\geq0.
\end{equation*}
Applying this bound frequency by frequency yields
\begin{equation}
\|W(\mathbf{y},\mathbf{k}_2)-W(\mathbf{y},\mathbf{k}_1)\|_2
\leq \frac{\|\mathbf{y}\|_2}{C}\|K_2-K_1\|_\infty
\leq \frac{\|\mathbf{y}\|_2}{C}
\|\mathbf{k}_2-\mathbf{k}_1\|_1.
\label{eq:supp_kernel_sensitivity}
\end{equation}
The last inequality follows from the triangle inequality for the Fourier sum. Finally, split the total difference into an observation change at $\mathbf{k}_2$ and a kernel change at $\mathbf{y}$, and combine the two bounds.
\end{proof}

The two terms in~\eqref{eq:supp_wiener_sensitivity} separate errors in the observation from errors in the estimated kernel. The constant $C$ prevents division by zero and limits sensitivity to both sources of error. Increasing $C$ reduces these bounds but can increase reconstruction bias. The gain in~\eqref{eq:supp_wiener_gain} may exceed one, so this result does not imply that Wiener deconvolution always suppresses noise. The kernel term allows an error in the estimated kernel to change the guidance by up to $\|\mathbf{y}\|_2/C$ times its $\ell_1$ norm, which is why the reliability of this estimate matters for injection.

Proposition~\ref{prop:supp_wiener_sensitivity} concerns the raw Wiener estimate. The subsequent NLM step in DAWN is not accounted for in Prop.~\ref{prop:supp_wiener_sensitivity}, and its effect depends on the filter strength $h$ (Table~\ref{tab:abl_nlm}). 

\subsection{Neural Refinement and the Simplified Residual}
\label{sec:supp_residual_scope}

Following~\eqref{eq:self_residual}, let $\mathbf{A}_{\mathbf{k}}$ denote circular convolution with $\mathbf{k}$ on the image grid, whose Fourier multiplier is $K$, and define the simplified residual
\begin{equation}
R(\mathbf{k})
:=\|\mathbf{A}_{\mathbf{k}}W(\mathbf{y},\mathbf{k})-\mathbf{y}\|_2.
\label{eq:supp_residual_definition}
\end{equation}
For nonnegative unit-sum kernels, Sec.~\ref{sec:hard_injection} shows that the Dirac delta kernel minimizes $R$, and the same holds for every shifted delta. Equality in~\eqref{eq:self_residual} only requires $|K(\omega)|=1$ at frequencies where $Y(\omega)\neq0$, so this minimizer need not be unique for a particular observation. With a softmax parameterization, which DAWN uses, a delta kernel can be approached but never obtained exactly.

The image produced by DAWN generally differs from $W(\mathbf{y},\mathbf{k})$. The following bound quantifies how much this difference can change the residual.

\begin{proposition}[Residual discrepancy for an actual image estimate]
\label{prop:supp_residual_discrepancy}
Let $\hat{\mathbf{k}}$ be nonnegative with unit sum. For any image estimate $\hat{\mathbf{x}}$, we have
\begin{equation}
\begin{aligned}
&\left|
\|\mathbf{A}_{\hat{\mathbf{k}}}\hat{\mathbf{x}}-\mathbf{y}\|_2
-R(\hat{\mathbf{k}})
\right|\leq
\|\hat{\mathbf{x}}-W(\mathbf{y},\hat{\mathbf{k}})\|_2.
\end{aligned}
\label{eq:supp_residual_discrepancy}
\end{equation}
Suppose $\hat{\mathbf{k}}_{\mathrm{inj}}$ is the kernel used at the most recent injection, and let
$\mathbf{z}_{\mathrm{inj}}=\mathcal{D}(W(\mathbf{y},\hat{\mathbf{k}}_{\mathrm{inj}}))$.
For the current image input $\mathbf{z}_x$, the right-hand side is bounded by
\begin{equation}
\|\hat{\mathbf{x}}-W(\mathbf{y},\hat{\mathbf{k}})\|_2
\leq
{\underbrace{\|\hat{\mathbf{x}}-\mathbf{z}_x\|_2}
_{\substack{\text{output--input}\\\text{mismatch}}}}
+
{\underbrace{\|\mathbf{z}_x-\mathbf{z}_{\mathrm{inj}}\|_2}
_{\substack{\text{input change}\\\text{since injection}}}}
+
{\underbrace{
\|\mathbf{z}_{\mathrm{inj}}
-W(\mathbf{y},\hat{\mathbf{k}}_{\mathrm{inj}})\|_2
}_{\text{NLM modification}}}
+
{\underbrace{
\frac{\|\mathbf{y}\|_2}{C}
\|\hat{\mathbf{k}}-\hat{\mathbf{k}}_{\mathrm{inj}}\|_1
}_{\substack{\text{effect of kernel change}\\\text{since injection}}}}.
\label{eq:supp_injection_discrepancy}
\end{equation}

\end{proposition}

\begin{proof}
Nonnegativity and unit mass imply $|\hat K(\omega)|\leq1$, hence
$\max_{\|\mathbf{x}\|_2 = 1} \|\mathbf{A}_{\hat{\mathbf{k}}}\mathbf{x}\|_2\leq1$.
The reverse triangle inequality gives
\begin{equation*}
\left|
\|\mathbf{A}_{\hat{\mathbf{k}}}\hat{\mathbf{x}}-\mathbf{y}\|_2
-R(\hat{\mathbf{k}})
\right|
\leq
\|\mathbf{A}_{\hat{\mathbf{k}}}
(\hat{\mathbf{x}}-W(\mathbf{y},\hat{\mathbf{k}}))\|_2
\leq
\|\hat{\mathbf{x}}-W(\mathbf{y},\hat{\mathbf{k}})\|_2.
\end{equation*}
For~\eqref{eq:supp_injection_discrepancy}, insert
$\mathbf{z}_x$, $\mathbf{z}_{\mathrm{inj}}$, and
$W(\mathbf{y},\hat{\mathbf{k}}_{\mathrm{inj}})$ between the two estimates, apply the triangle inequality, and bound the remaining Wiener difference using~\eqref{eq:supp_kernel_sensitivity}.
\end{proof}

The four terms on the right-hand side of \eqref{eq:supp_injection_discrepancy} describe the output--input mismatch, changes to the input since injection, the modification introduced by NLM, and changes to the kernel since injection. When all four are small, the simplified residual is close to the current image's residual.

The bound clarifies the role of neural refinement. Under per-step injection, the input is reset to the injected estimate at every outer iteration, so the second term vanishes and the fourth covers a single outer iteration. The actual residual then stays within the network and NLM terms of the simplified residual, which favors the no-blur solution. Under alternation, neural refinement replaces the input with the network's output between injections, so the second and fourth terms can grow and the bound no longer ties the actual residual to the simplified one. The data-fidelity term can then constrain the kernel again, as discussed in Sec.~\ref{sec:hard_injection}.

These terms are not guaranteed to be small, even right after an injection, because the NLM step is designed to remove the noise that Wiener deconvolution amplifies and can therefore modify the raw estimate substantially. The calculation in~\eqref{eq:self_residual}, where the delta kernel attains the lower bound, thus gives a simplified argument that applies to per-step injection of the raw Wiener estimate. For the full pipeline, with the network and the denoiser included, the collapse to near-delta kernels under per-step injection is an empirical finding (Table~\ref{tab:kernel_collapse}).

\section{Evaluation Metrics}
\label{sec:supp_metrics}

We report PSNR and SSIM~\citep{wang2004ssim} for image reconstruction quality. For kernel estimation, we use the Maximum Normalized Correlation (MNC):
\begin{equation}
  \mathrm{MNC}(\hat{\mathbf{k}}, \mathbf{k}^*) = \max_{(i,j)} \frac{(\hat{\mathbf{k}} \star \mathbf{k}^*)(i,j)}{\|\hat{\mathbf{k}}\|_2 \cdot \|\mathbf{k}^*\|_2},
\end{equation}
where $\star$ denotes 2D cross-correlation, $\hat{\mathbf{k}}$ is the estimated kernel, and $\mathbf{k}^*$ is the ground truth.
$\mathrm{MNC} \in [0, 1]$, with $\mathrm{MNC} = 1$ indicating a perfect match up to translation~\citep{levin2009understanding}.

For the microscopy dataset, we additionally report the DICE coefficient to measure downstream segmentation fidelity. We first binarize both the deblurred output and the ground-truth image using Otsu thresholding, yielding binary segmentation masks $A$ and $B$, respectively. The DICE coefficient is then computed as
\begin{equation}
  \mathrm{DICE}(A, B) = \frac{2\,|A \cap B|}{|A| + |B|},
\end{equation}
where $|\cdot|$ denotes the number of foreground pixels. $\mathrm{DICE} \in [0, 1]$, with $\mathrm{DICE} = 1$ indicating perfect segmentation overlap. All metrics are computed after shift-alignment between the estimated and ground-truth images.

\section{Implementation Details}
\label{sec:supp_implementation}

DAWN uses the same U-Net image generator and FCN kernel generator architectures as SelfDeblur~\cite{ren2020selfdeblur}. DAWN typically converges by around 4000 gradient steps, divided into $K = 2000$ outer iterations of $N = 2$ inner steps each.
The Wiener injection occurs every $M = 200$ outer iterations (i.e., every 400 gradient steps) starting from the first iteration.
We use Adam~\citep{kingma2015adam} with learning rates $10^{-3}$ (image network) and $10^{-5}$ (kernel network).
The autoencoding weights are $\lambda_x = 1.0$ (image) and $\lambda_k = 2.0$ (kernel), with kernel EMA momentum $\gamma = 0.5$ and Wiener regularization constant $C = 0.05$.
Input regularization noise has standard deviation $\sigma_z = 0.01$.
The non-local means denoiser $\mathcal{D}$ follows the pipeline of \citet{zhong2013cvpr} with filter strength $h = 20$.
All neural methods are run on NVIDIA A40 GPUs; the classical baselines are run on CPU using their MATLAB implementations.

Descriptions and configurations of all baselines are provided in Sec.~\ref{sec:supp_baselines}. All results are obtained from a single run per image with independent random seeds.

\section{Baseline Methods}
\label{sec:supp_baselines}

We compare against baselines spanning DIP-based neural optimization, diffusion-based posterior sampling, and classical MAP estimation. We describe each method and its configuration below. For all baselines we used the official implementation with its provided default hyperparameters.

\textbf{SelfDeblur~\cite{ren2020selfdeblur}.}
SelfDeblur uses the Deep Image Prior framework for blind deconvolution, jointly optimizing a U-Net image generator and a softmax-constrained FCN kernel generator on the blurry observation. 

\textbf{W-DIP~\cite{bredell2023wdip}.}
W-DIP augments the SelfDeblur formulation with Wiener guidance: an auxiliary kernel, initialized as a Gaussian and coupled to the kernel network through half-quadratic splitting, produces a Wiener-deconvolved (sharpened) version of the blurry image, and additional penalty terms pull the image generator toward this deconvolved guide and the kernel network toward the auxiliary kernel during optimization. Like SelfDeblur, W-DIP switches from an MSE to an SSIM loss at gradient step 1000. 

\textbf{\citet{lee2024eccv}.}
This method derives a noise-robust kernel estimation function that computes the blur kernel analytically from a clear image and the blurry measurement: in the Fourier domain, $\widehat{H} = \overline{X}\,Y / (|X|^2 + k)$, where $X$ and $Y$ denote the Fourier transforms of the clear-image estimate and the blurry image, and the noise suppression factor $k$ (optimally proportional to the noise variance) provides robustness to noise. A DIP generator supplies the clear-image estimate and is optimized with reblurring data-fidelity losses in which the kernel is given by this estimator, evaluated under multiple suppression factors to cover unknown noise levels. 

\textbf{GibbsDDRM~\cite{murata2023gibbsddrm}.}
GibbsDDRM extends Denoising Diffusion Restoration Models (DDRM) to blind linear inverse problems: it forms a joint distribution over the image, the measurement, and the blur kernel, using a pre-trained diffusion model as the image prior and a simple generic (Laplace) prior on the kernel, and performs posterior sampling with a partially collapsed Gibbs sampler that alternates DDRM-style updates of the image's diffusion latents, conditioned on the current kernel estimate, with approximate sampling of the kernel from its conditional posterior. We used the official implementation with its FFHQ $256 \times 256$ pre-trained diffusion model.

\textbf{\citet{krishnan2011blind}.}
This method employs a normalized sparsity measure, the $\ell_1/\ell_2$ ratio of image gradients, as a scale-invariant regularization function to prevent the diminishment of high-frequency components during the coarse-to-fine kernel estimation process. 

\textbf{Pan-DCP~\cite{pan2016blind}.}
This method leverages the dark channel prior of natural images, the observation that most local patches contain pixels with very low intensity in at least one color channel, and incorporates it as a regularization function for blind deconvolution. We used the official implementation with default hyperparameters: $\lambda_{\mathrm{dark}} = 4 \times 10^{-3}$, $\lambda_{\mathrm{grad}} = 4 \times 10^{-3}$, $\lambda_{\mathrm{TV}} = 3 \times 10^{-3}$, and $\lambda_{\ell_0} = 5 \times 10^{-4}$.

\textbf{\citet{dong2017outlier}.}
This is a classical MAP-based blind deblurring method that replaces the standard $\ell_2$ data-fidelity term with a robust function $\mathcal{R}(z)$ that saturates for large residuals, effectively down-weighting outlier pixels during kernel estimation without explicit outlier detection. The method uses a coarse-to-fine IRLS optimization. 

Baselines excluded from the comparison, and why, are discussed in Sec.~\ref{sec:experiments}.

\section{Additional Experimental Results}
\label{sec:supp_additional}

\begin{figure}[t]
\centering
\includegraphics[width=\textwidth]{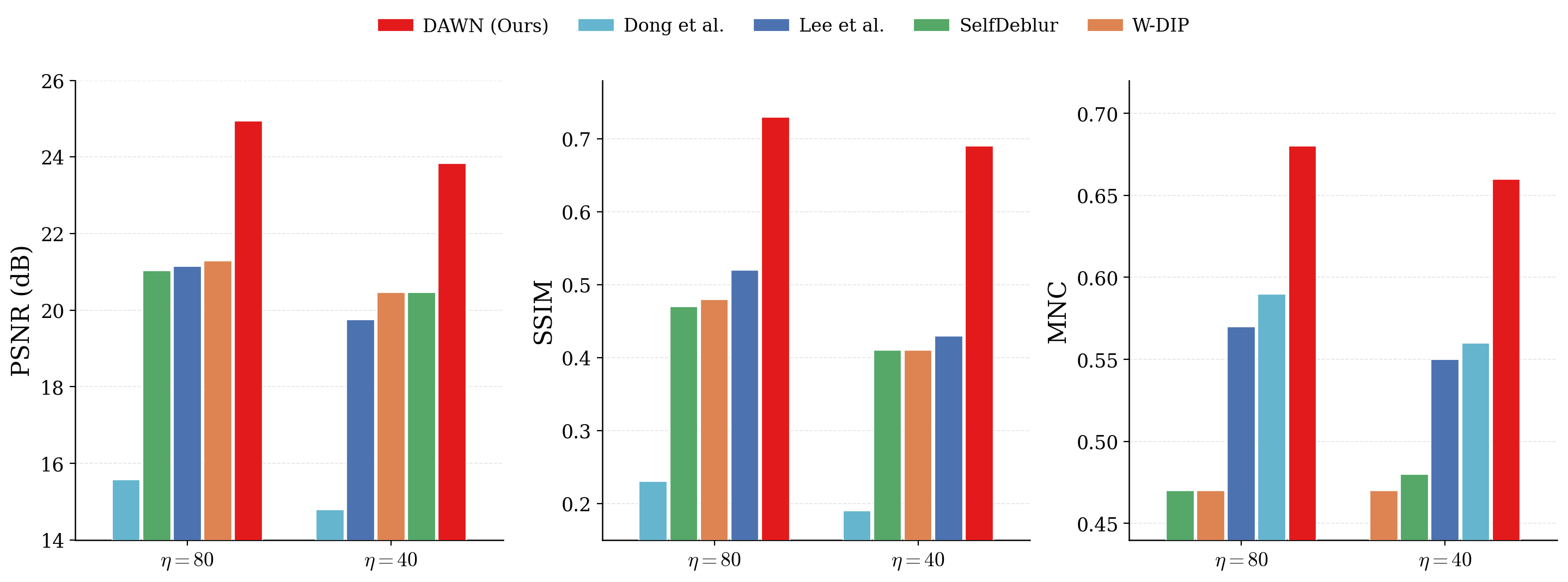}
\caption{Poisson shot noise results on Levin~\citep{levin2009understanding} at $\eta = 80$ and $\eta = 40$. Bars sorted in ascending order. DAWN uses the same hyperparameters as the Gaussian experiments.}
\label{fig:shot_noise}
\end{figure}

\textbf{Robustness to shot noise.}
To assess robustness to non-Gaussian noise, we evaluate on Poisson shot noise following \citet{zhuang2024ijcv}, where $\mathbf{y} = \mathrm{Poisson}(\eta \cdot \mathbf{k} \ast \mathbf{x}) / \eta$ and higher $\eta$ corresponds to weaker noise.
Fig.~\ref{fig:shot_noise} shows results on Levin at $\eta = 80$ (moderate) and $\eta = 40$ (severe): DAWN outperforms all baselines in PSNR, SSIM, and MNC at both levels using identical hyperparameters, demonstrating robust transfer to non-Gaussian noise.

\textbf{Real-world images.}
Figs.~\ref{fig:supp_real_flower}--\ref{fig:supp_real_text10}, at the end of the appendix, show four additional real-world images from \citet{lai2016comparative} with noise of $\sigma = 0.05$. As no ground truth is available, all methods use a $51 \times 51$ kernel support. DAWN produces clearer images and better kernel estimates than all baselines.

\textbf{Robustness to kernel size overspecification.}
So far, every method, including DAWN, has used the ground-truth kernel size as its kernel support. As the kernel size is rarely known in real-world imaging, where the support is usually overspecified, we enlarge the support beyond the ground-truth kernel size by 4, 8, 16, and 24 pixels in each dimension on the first 8 images of Levin at $\sigma = 0.05$, which cover all eight kernels (13 to 27 pixels wide), with all hyperparameters fixed. As Table~\ref{tab:kernel_oversize} shows, the drop in PSNR is very modest, even when the kernel support is overspecified to nearly three times the ground-truth size, and the kernel MNC falls very modestly. This shows that DAWN is not very sensitive to overspecified kernel size, making it applicable in practical imaging settings where the blur and its size is unknown.

\begin{table}[h]
\centering
\caption{Results of DAWN on the first 8 Levin images ($\sigma = 0.05$) with the kernel support enlarged by $n$ pixels per dimension ($+n$). The width ratio is the support width over the ground-truth kernel width.}
\label{tab:kernel_oversize}
\setlength{\tabcolsep}{4pt}
\small
\begin{tabular}{l | cccc}
\toprule
Kernel support & Width ratio & PSNR$\uparrow$ & SSIM$\uparrow$ & MNC$\uparrow$ \\
\midrule
\textbf{Ground-truth size} & \textbf{1.0} & \textbf{26.16} & \textbf{0.77} & \textbf{0.73} \\
$+4$ & 1.1--1.3 & 26.03 & 0.77 & 0.73 \\
$+8$ & 1.3--1.6 & 25.40 & 0.76 & 0.72 \\
$+16$ & 1.6--2.2 & 25.67 & 0.76 & 0.71 \\
$+24$ & 1.9--2.8 & 25.44 & 0.75 & 0.70 \\
\bottomrule
\end{tabular}
\end{table}

\textbf{Runtime.}
Table~\ref{tab:runtime} reports the average per-sample runtime on FFHQ ($256 \times 256$). Among the DIP-based methods, DAWN is faster than both SelfDeblur and W-DIP. Timings are not directly comparable across method families: the neural methods are run on NVIDIA A40 GPUs, whereas the classical MAP baselines are run on CPU using their MATLAB implementations.

\begin{table}[h]
\centering
\caption{Average per-sample runtime on FFHQ ($256 \times 256$).}
\label{tab:runtime}
\setlength{\tabcolsep}{6pt}
\small
\begin{tabular}{l | c}
\toprule
Method & Runtime (s)$\downarrow$ \\
\midrule
\multicolumn{2}{l}{\emph{DIP-based (training-data-free)}} \\
\textbf{DAWN (Ours)} & 96.3 \\
SelfDeblur & 118.6 \\
W-DIP & 167.6 \\
Lee et al. & 48.5 \\
\midrule
\multicolumn{2}{l}{\emph{Diffusion prior (pre-trained)}} \\
GibbsDDRM & 69.1 \\
\midrule
\multicolumn{2}{l}{\emph{Classical MAP}} \\
Krishnan et al. & 4.8 \\
Pan-DCP & 97.4 \\
Dong et al. & 59.7 \\
\bottomrule
\end{tabular}
\end{table}

\section{Ablation Studies}
\label{sec:supp_ablation}

All ablations are conducted on the Levin dataset due to its smaller size. Table~\ref{tab:ablation} reports the effect of removing each major component; the remaining ablations vary individual hyperparameters.

\begin{table}[h]
\centering
\caption{Component ablation on Levin. Each row removes one component from the full method.}
\label{tab:ablation}
\setlength{\tabcolsep}{4pt}
\small
\begin{tabular}{l | ccc}
\toprule
Configuration & PSNR$\uparrow$ & SSIM$\uparrow$ & MNC$\uparrow$ \\
\midrule
\textbf{DAWN (full)} & \textbf{25.02} & \textbf{0.75} & \textbf{0.70} \\
\quad w/o Wiener injection & 21.15 & 0.59 & 0.70 \\
\quad w/o kernel AE ($\lambda_k = 0$) & 24.55 & 0.71 & 0.65 \\
\quad w/o image AE ($\lambda_x = 0$) & 24.07 & 0.70 & 0.70 \\
\bottomrule
\end{tabular}
\end{table}

\begin{table}[h]
\centering
\caption{Ablation on the image autoencoding weight $\lambda_x$ (left) and the Wiener injection period $M$ (right), both on Levin.}
\label{tab:abl_lx_M}
\setlength{\tabcolsep}{2.5pt}
\small
\begin{tabular}[t]{c | ccc}
\toprule
$\lambda_x$ & PSNR$\uparrow$ & SSIM$\uparrow$ & MNC$\uparrow$ \\
\midrule
0.5 & 24.39 & 0.71 & 0.68 \\
\textbf{1.0} & \textbf{25.02} & \textbf{0.75} & \textbf{0.70} \\
2.0 & 24.97 & 0.73 & 0.69 \\
3.0 & 24.82 & 0.72 & 0.66 \\
\bottomrule
\end{tabular}
\qquad
\begin{tabular}[t]{c | ccc}
\toprule
$M$ & PSNR$\uparrow$ & SSIM$\uparrow$ & MNC$\uparrow$ \\
\midrule
1   & 22.59 & 0.65 & 0.49 \\
50  & 24.08 & 0.70 & 0.63 \\
100 & 24.60 & 0.71 & 0.66 \\
\textbf{200} & \textbf{25.02} & \textbf{0.75} & \textbf{0.70} \\
300 & 24.52 & 0.72 & 0.69 \\
400 & 24.41 & 0.70 & 0.68 \\
\bottomrule
\end{tabular}
\end{table}

\begin{table}[h]
\centering
\caption{Ablation on the NLM denoiser strength $h$ on Levin ($\sigma = 0.05$). $h = 0$ disables denoising.}
\label{tab:abl_nlm}
\setlength{\tabcolsep}{4pt}
\small
\begin{tabular}{c | ccc}
\toprule
$h$ & PSNR$\uparrow$ & SSIM$\uparrow$ & MNC$\uparrow$ \\
\midrule
0  & 24.61 & 0.71 & 0.67 \\
10 & 24.87 & 0.72 & 0.68 \\
\textbf{20} & \textbf{25.02} & \textbf{0.75} & \textbf{0.70} \\
30 & 24.39 & 0.70 & 0.67 \\
\bottomrule
\end{tabular}
\end{table}

\begin{table}[h]
\centering
\caption{Mean effective number of pixels of the kernels on Levin ($\sigma = 0.05$).}
\label{tab:kernel_collapse}
\setlength{\tabcolsep}{6pt}
\small
\begin{tabular}{l | c}
\toprule
Kernels & Effective number of pixels \\
\midrule
DAWN ($M = 200$) & 63.8 \\
$M = 1$ (every iteration) & 1.3 \\
\bottomrule
\end{tabular}
\end{table}

\textbf{Image autoencoding weight $\lambda_x$.}
Table~\ref{tab:abl_lx_M} (left) varies $\lambda_x$ with all other hyperparameters fixed. Performance is robust across the range, with $\lambda_x = 1.0$ yielding the best PSNR, SSIM, and MNC. The default choice $\lambda_x = 1.0$ inherits from aSeqDIP's optimal choice of $\lambda$.

\textbf{Wiener injection period $M$.}
Table~\ref{tab:abl_lx_M} (right) varies the injection period $M$ (in outer iterations). Injecting at every outer iteration ($M = 1$) reduces DAWN to direct injection at every iteration (Sec.~\ref{sec:hard_injection}), which collapses the estimated kernels to near-deltas and removes most of the benefit of injection.
Table~\ref{tab:kernel_collapse} quantifies this collapse with the effective number of pixels, defined by $1/\sum_i k_i^2$, where $k_i$ denotes the $i$-th entry of the kernel $\mathbf{k}$. It equals 1 for a delta and $n$ for a uniform kernel over $n$ pixels. With $M = 1$, the effective number of pixels is at most 1.4 on all 32 images, consistent with the simplified argument in Sec.~\ref{sec:hard_injection}, whereas DAWN's kernels do not collapse. Frequent injection (e.g., $M = 50$) leaves fewer refinement steps between injections and degrades performance, while infrequent injection (e.g., $M \geq 300$) reduces the number of corrections within the optimization budget.

\textbf{NLM denoiser strength $h$.}
Table~\ref{tab:abl_nlm} varies the filter strength $h$ of the NLM denoiser $\mathcal{D}$ (on the 8-bit intensity scale) on Levin at $\sigma = 0.05$. With $h = 0$ the denoiser is disabled, so each injection resets $\mathbf{z}_x$ to the raw Wiener estimate $W(\mathbf{y}, \hat{\mathbf{k}})$.
A moderate strength ($h = 10$ or $20$) improves PSNR, SSIM, and MNC over $h = 0$, as NLM suppresses the noise that Wiener deconvolution amplifies before it reaches the network input. Too strong a filter ($h = 30$) removes image detail, as it is worse than the default on all three metrics and falls below $h = 0$ in PSNR and SSIM.
Even without denoising, DAWN reaches 24.61 dB, well above every baseline in Table~\ref{tab:gaussian} (at most 21.96 dB) and above the 21.15 dB obtained without Wiener injection (Table~\ref{tab:ablation}). Most of DAWN's gain over the baselines therefore comes from the direct injection of the Wiener estimate into the network input, with NLM denoising as a secondary improvement.

\textbf{Kernel autoencoding hyperparameters $\lambda_k$ and $\gamma$.}
These were selected via grid search on the first 8 images of the Levin dataset, yielding $\lambda_k = 2.0$ and $\gamma = 0.5$.

\clearpage

\begin{figure}[p]
\centering
\includegraphics[width=\textwidth]{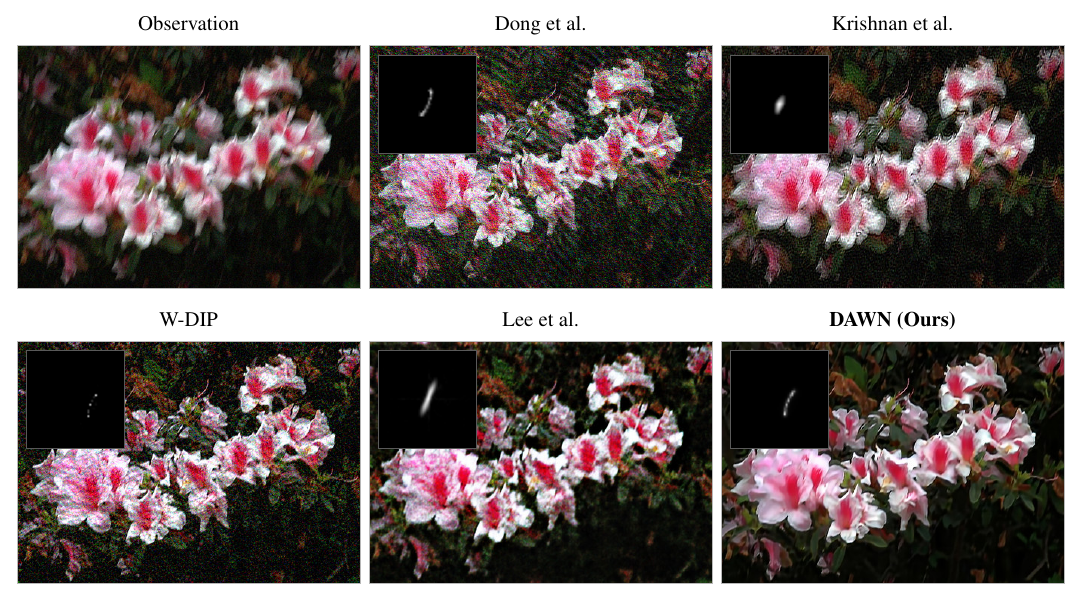}
\caption{Qualitative comparison on a real-world blurry image from the real-world subset of \citet{lai2016comparative}, with noise of $\sigma = 0.05$. No ground truth image or kernel is available, and all methods use a $51 \times 51$ kernel support. Zoom in for best viewing.}
\label{fig:supp_real_flower}
\end{figure}
\clearpage

\begin{figure}[p]
\centering
\includegraphics[width=\textwidth]{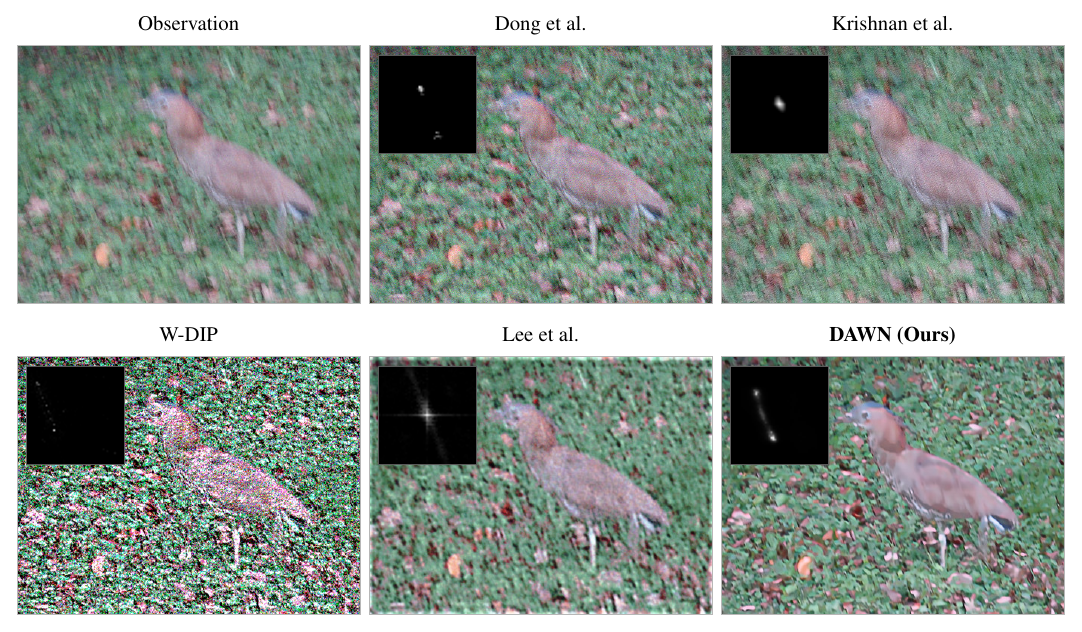}
\caption{Qualitative comparison on a second real-world blurry image from the real-world subset of \citet{lai2016comparative}, with noise of $\sigma = 0.05$. No ground truth image or kernel is available, and all methods use a $51 \times 51$ kernel support. Zoom in for best viewing.}
\label{fig:supp_real_bird}
\end{figure}
\clearpage

\begin{figure}[p]
\centering
\includegraphics[width=\textwidth]{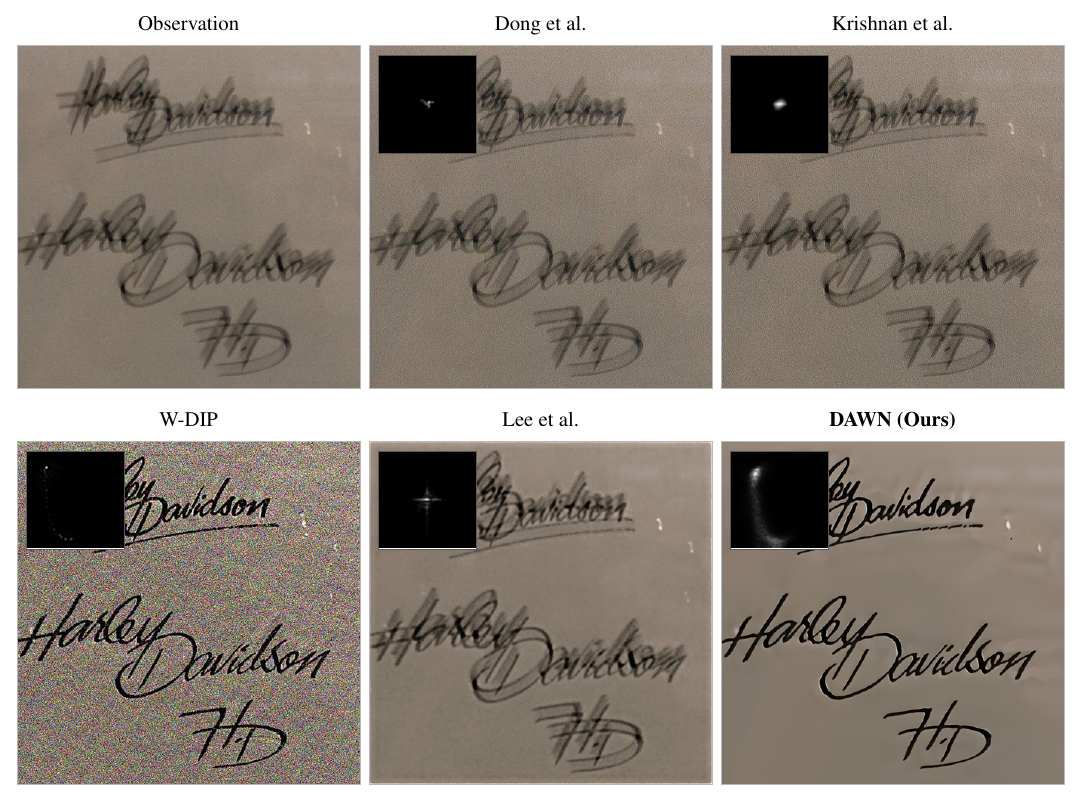}
\caption{Qualitative comparison on a third real-world blurry image from the real-world subset of \citet{lai2016comparative}, with noise of $\sigma = 0.05$. No ground truth image or kernel is available, and all methods use a $51 \times 51$ kernel support. Zoom in for best viewing.}
\label{fig:supp_real_text2}
\end{figure}
\clearpage

\begin{figure}[p]
\centering
\includegraphics[width=\textwidth]{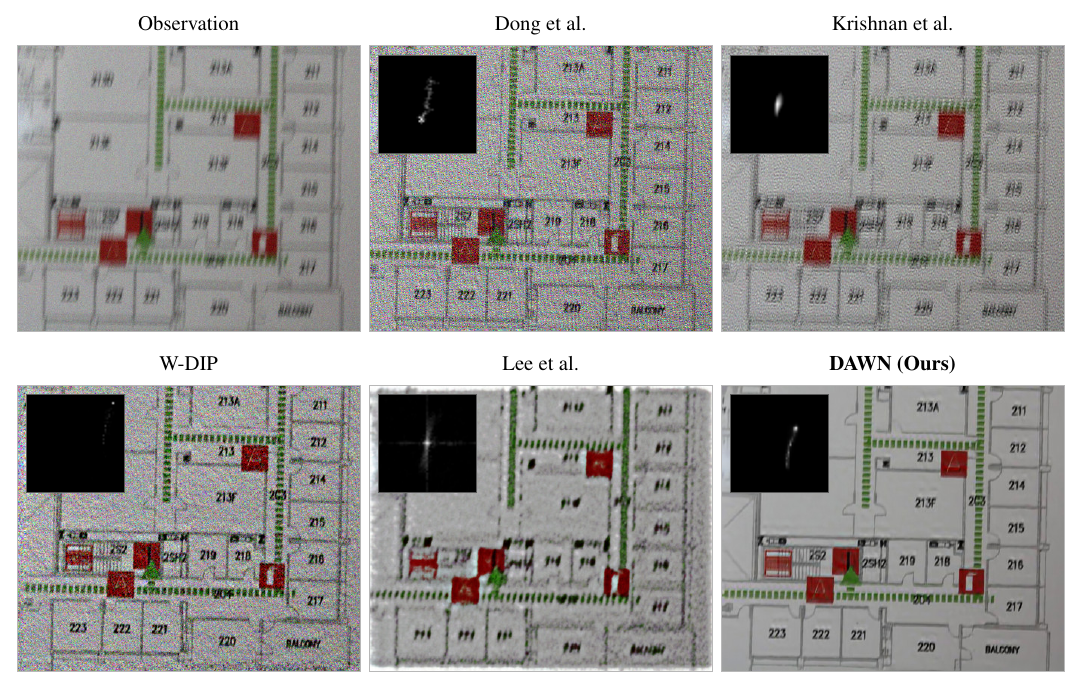}
\caption{Qualitative comparison on a fourth real-world blurry image from the real-world subset of \citet{lai2016comparative}, with noise of $\sigma = 0.05$. No ground truth image or kernel is available, and all methods use a $51 \times 51$ kernel support. Zoom in for best viewing.}
\label{fig:supp_real_text10}
\end{figure}
\clearpage

\begin{figure}[p]
\centering
\includegraphics[width=\textwidth]{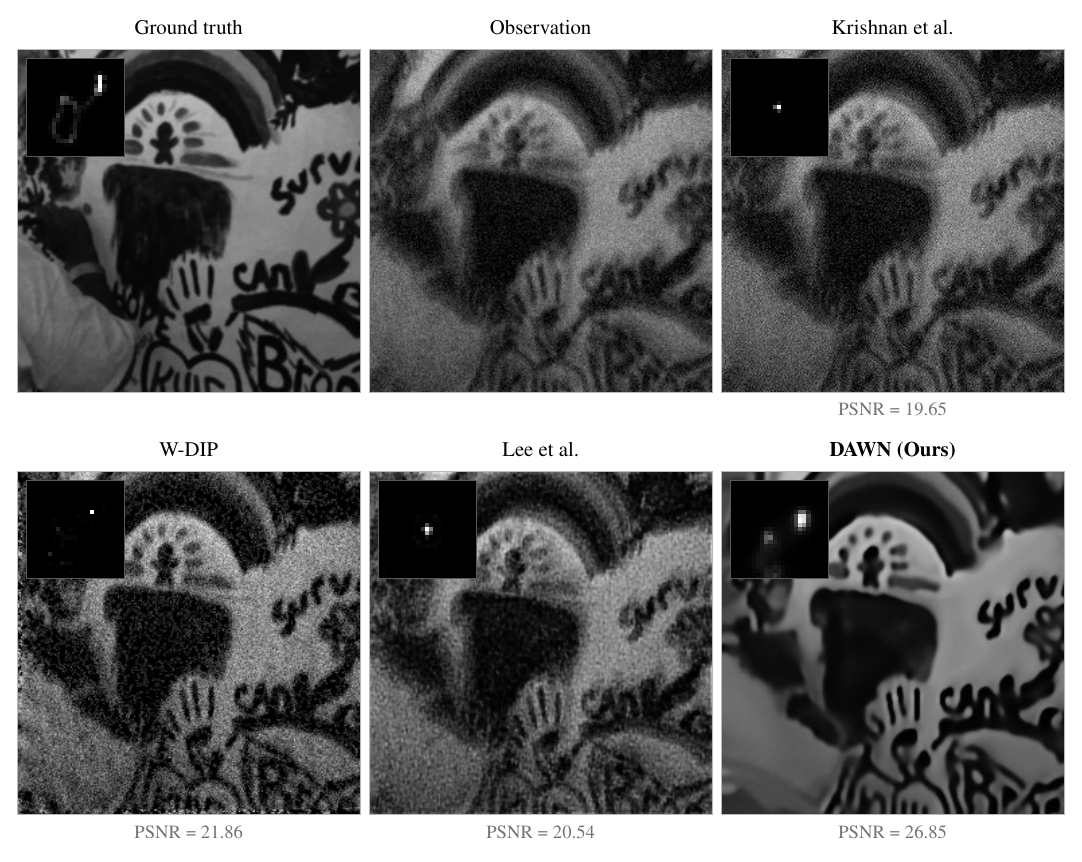}
\caption{Qualitative comparison on a Levin image at $\sigma = 0.1$. Zoom in for best viewing.}
\label{fig:qual_levin1}
\end{figure}

\begin{figure}[p]
\centering
\includegraphics[width=\textwidth]{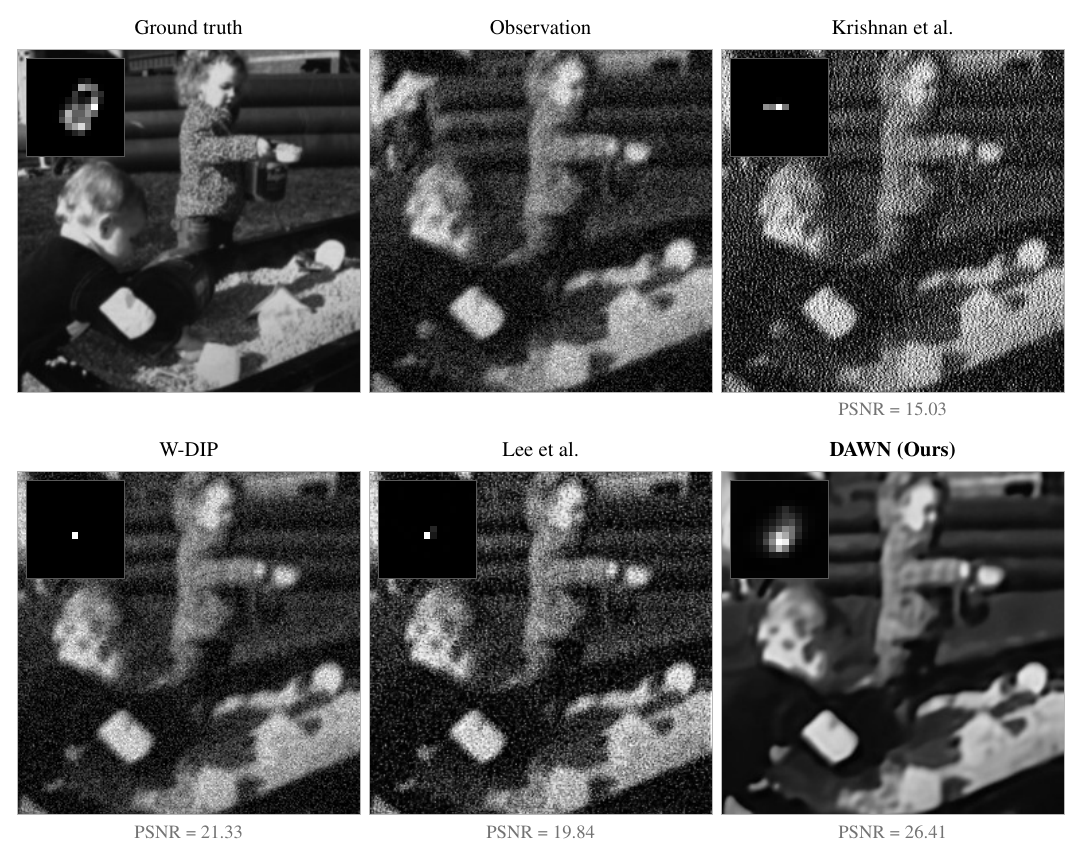}
\caption{Qualitative comparison on a second Levin image at $\sigma = 0.1$. Zoom in for best viewing.}
\label{fig:qual_levin2}
\end{figure}

\begin{figure}[p]
\centering
\includegraphics[width=\textwidth]{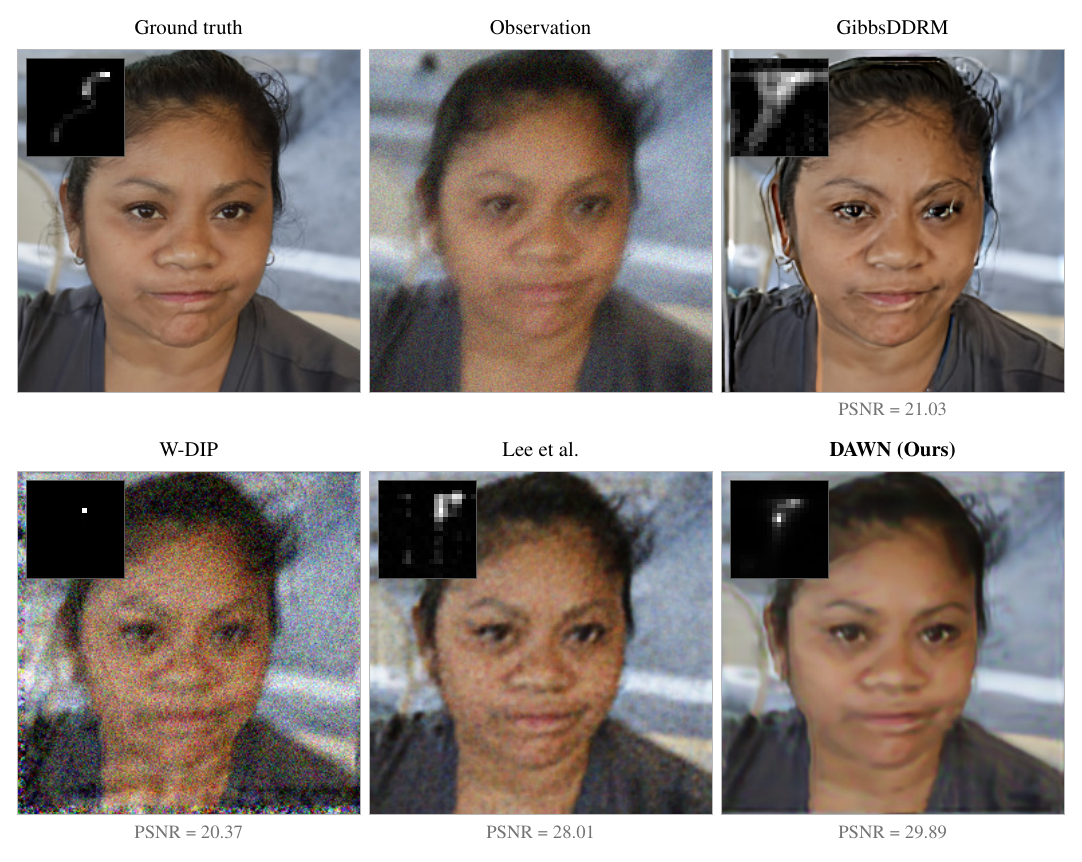}
\caption{Qualitative comparison on an FFHQ image at $\sigma = 0.05$. Zoom in for best viewing.}
\label{fig:qual_ffhq}
\end{figure}

\begin{figure}[p]
\centering
\includegraphics[width=\textwidth]{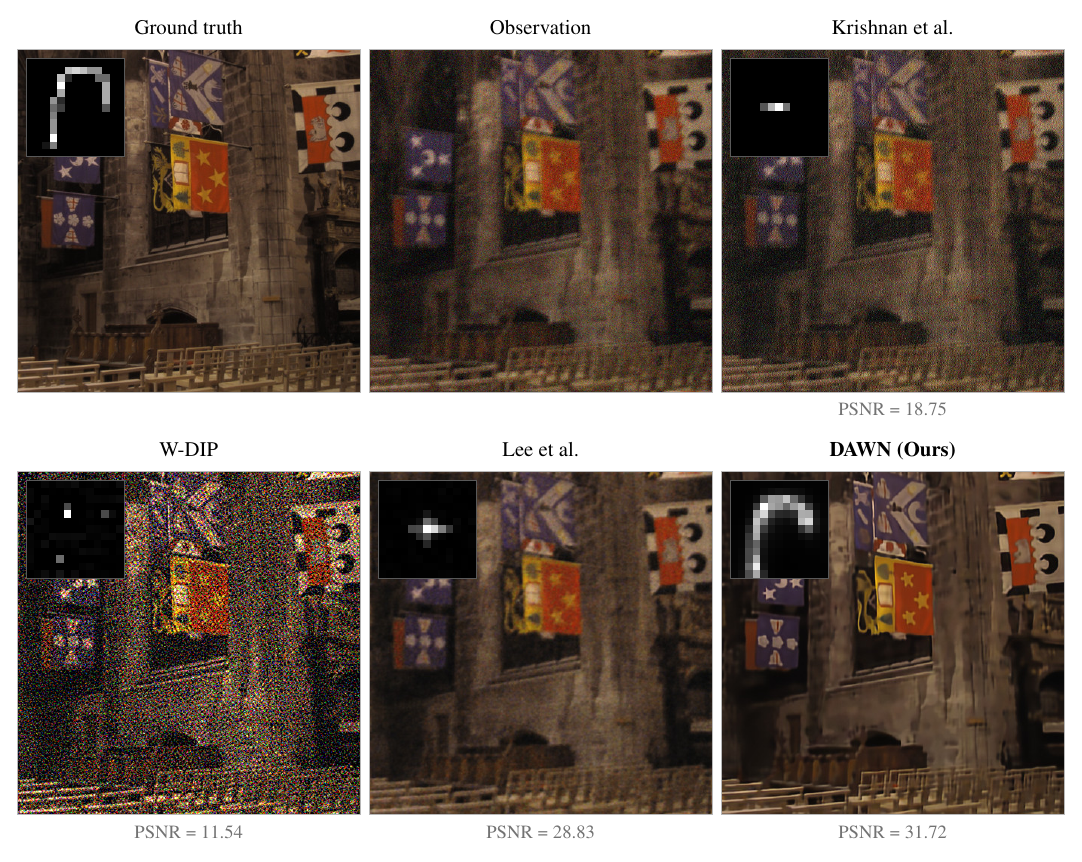}
\caption{Qualitative comparison on a K\"{o}hler image at $\sigma = 0.1$. Zoom in for best viewing.}
\label{fig:qual_kohler1}
\end{figure}

\begin{figure}[p]
\centering
\includegraphics[width=\textwidth]{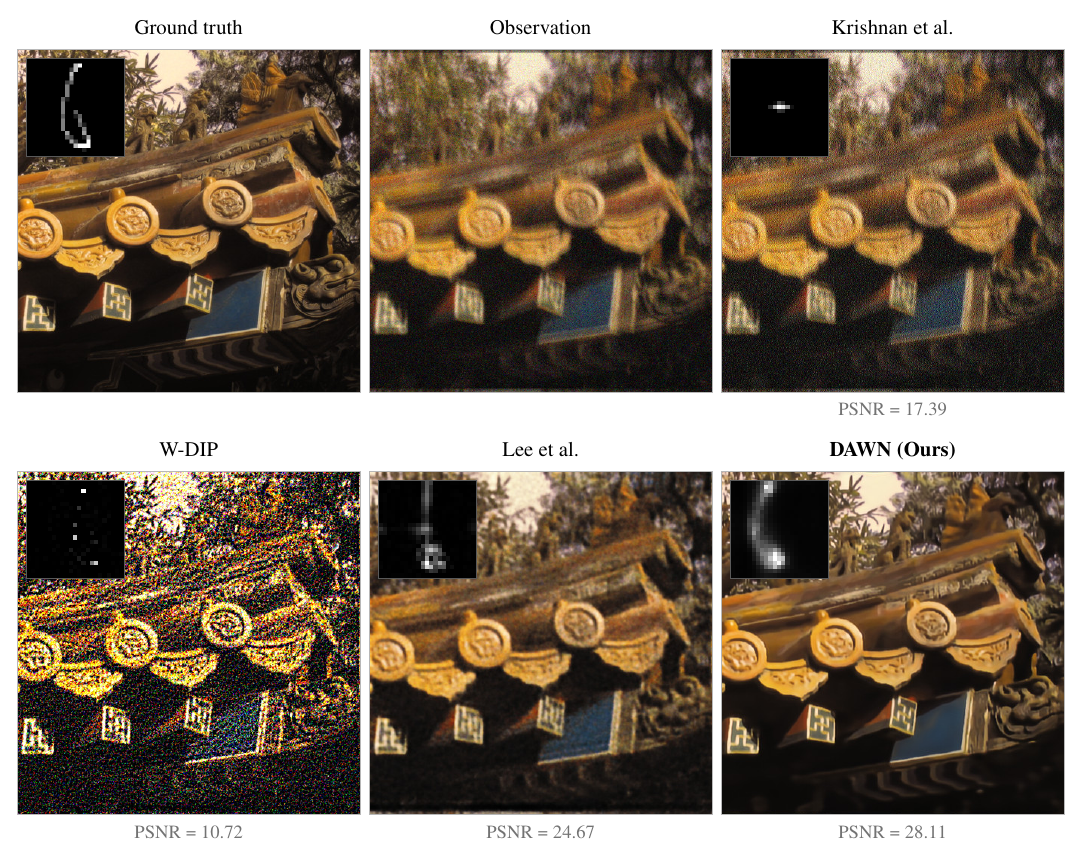}
\caption{Qualitative comparison on a second K\"{o}hler image at $\sigma = 0.1$. Zoom in for best viewing.}
\label{fig:qual_kohler2}
\end{figure}

\begin{figure}[p]
\centering
\includegraphics[width=\textwidth]{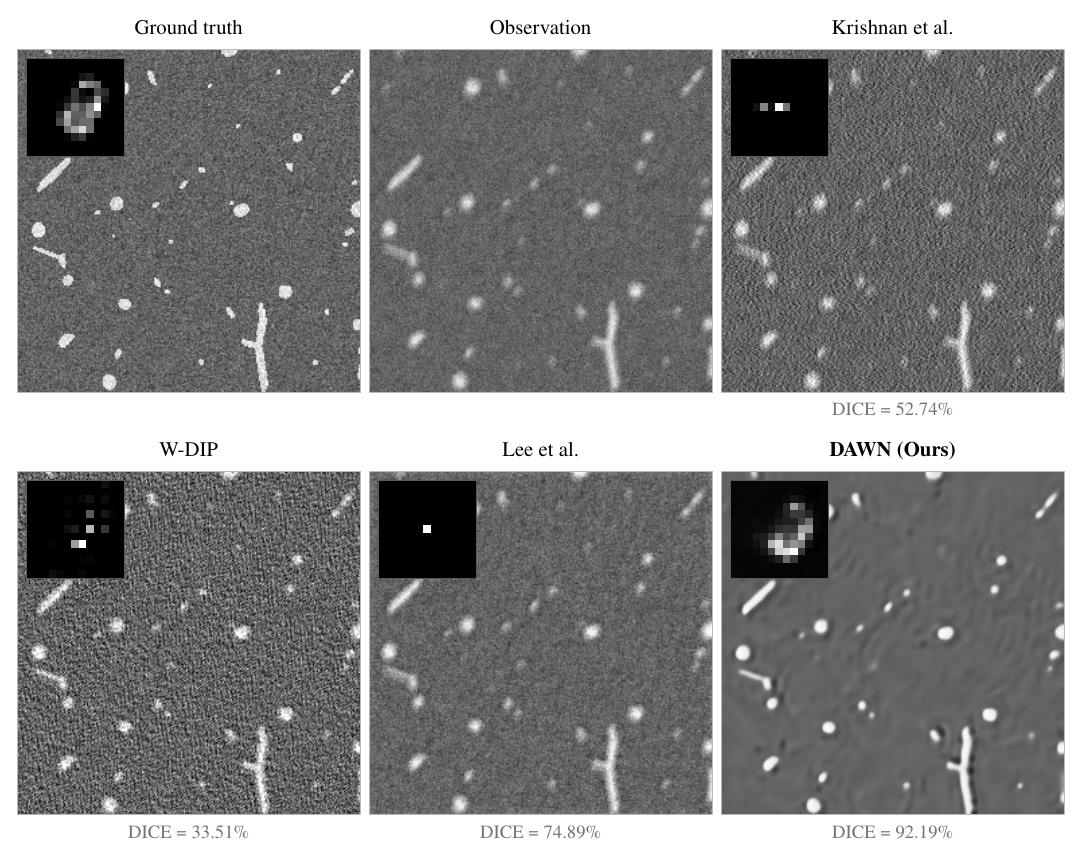}
\caption{Qualitative comparison on a microscopy image at $\sigma = 0.05$. Zoom in for best viewing.}
\label{fig:qual_micro}
\end{figure}

\typeout{get arXiv to do 4 passes: Label(s) may have changed. Rerun}
\end{document}